\documentclass[10pt,twocolumn,letterpaper]{article}

\usepackage[export]{adjustbox}
\usepackage{multirow}
\usepackage{threeparttable}
\usepackage{subcaption}
\usepackage{dirtytalk}
\usepackage{xcolor}
\usepackage{float}
\usepackage[switch]{lineno}
\usepackage[pagenumbers]{wacv} % To force page numbers, e.g. for an arXiv version
\usepackage[accsupp]{axessibility}  % Improves PDF readability for those with disabilities.

\definecolor{wacvblue}{rgb}{0.21,0.49,0.74}
\usepackage[pagebackref,breaklinks,colorlinks,allcolors=wacvblue]{hyperref}

\def\wacvPaperID{} % *** Enter the WACV Paper ID here
\def\confName{WACV}
\def\confYear{2026}

\title{CONSTANT: Towards High-Quality One-Shot Handwriting Generation with Patch \underline{Con}trastive Enhancement and \underline{St}yle-\underline{A}ware Qua\underline{nt}ization}

\author{Anh-Duy Le$^1$, Van-Linh Pham$^1$, Thanh-Nam Vo$^1$, Xuan Toan Mai$^2$, Tuan-Anh Tran$^{1,2}$\\
$^1$ Viettel Artificial Intelligence and Data Services Center, Vietnam \\
$^2$ Ho Chi Minh City University of Technology, Vietnam \\ 
{\tt\small \{leanhduy497, phamvanlinh143, thanhnam040501\}@gmail.com, \{mxtoan, trtanh\}@hcmut.edu.vn}
}

\begin{document}
\maketitle
\begin{abstract}
One-shot styled handwriting image generation, despite achieving impressive results in recent years, remains challenging due to the difficulty in capturing the intricate and diverse characteristics of human handwriting by using solely a single reference image. Existing methods still struggle to generate visually appealing and realistic handwritten images and adapt to complex, unseen writer styles, struggling to isolate invariant style features (e.g., slant, stroke width, curvature) while ignoring irrelevant noise. To tackle this problem, we introduce Patch \textbf{Con}trastive Enhancement and \textbf{St}yle-\textbf{A}ware Qua\textbf{nt}ization via Denoising Diffusion (\textbf{CONSTANT}), a novel one-shot handwriting generation via diffusion model. CONSTANT leverages three key innovations: 1) a Style-Aware Quantization (SAQ) module that models style as discrete visual tokens capturing distinct concepts; 2) a contrastive objective to ensure these tokens are well-separated and meaningful in the embedding style space; 3) a latent patch-based contrastive ($L_{LatentPCE}$) objective help improving quality and local structures by aligning multiscale spatial patches of generated and real features in latent space. Extensive experiments and analysis on benchmark datasets from multiple languages, including English, Chinese, and our proposed ViHTGen dataset for Vietnamese, demonstrate the superiority of adapting to new reference styles and producing highly detailed images of our method over state-of-the-art approaches. Code is available at \href{https://github.com/duylebkHCM/CONSTANT}{GitHub}
\end{abstract}
    
\section{Introduction}
\begin{figure}[t]
    \centering
    \begin{subfigure}[b]{0.8\linewidth}
        \includegraphics[width=\linewidth]{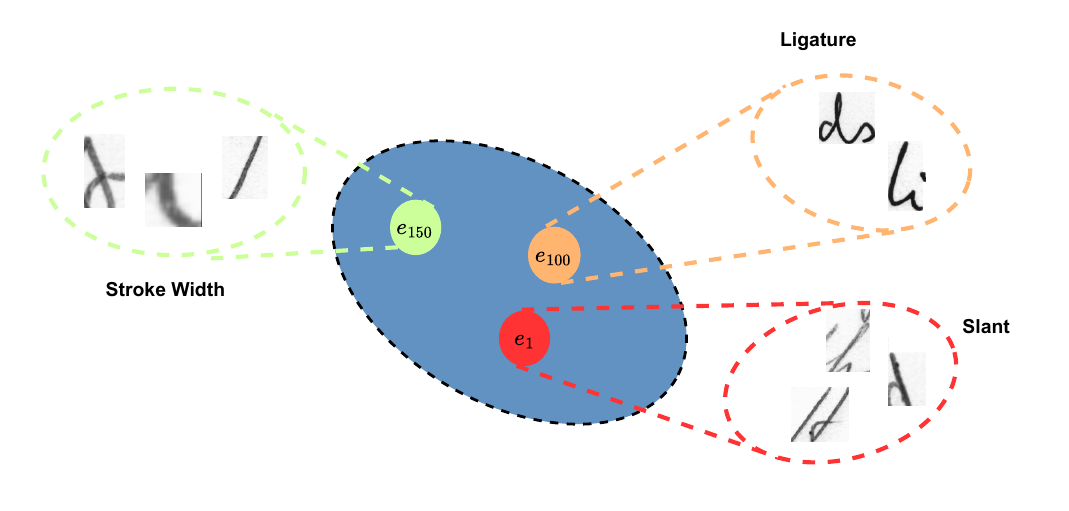}
        \caption{Handwritten style concepts in an embedding space}
        \label{fig:styleconcept}
    \end{subfigure}
    \vskip-0.25\baselineskip % Reduce vertical space here
    \begin{subfigure}[b]{0.8\linewidth}
        \includegraphics[width=\linewidth]{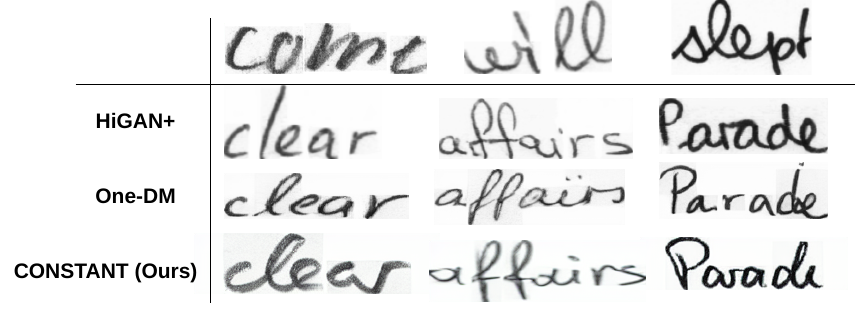}
        \caption{Style mimicking ability between our CONSTANT and other methods (One-DM~\cite{dai2025one}, HiGAN+~\cite{HiGANplus})}
        \label{fig:compstyle}
    \end{subfigure}
    \caption{Our method leverages the idea of vector quantization to learn the embedding space of diverse style concepts, enabling better adaptation to new styles than other approaches.}

    \label{fig:mainidea}
    \vspace{-0.2in}
\end{figure}

Handwriting synthesis is a significant problem with applications ranging from assistive technology to data augmentation for robust authentication~\cite{hong2024handwritten} and text recognition systems~\cite{ingle2019scalable}. The one-shot approach, which generates handwriting from a single style example, is crucial for practical applications where users may only provide one reference. However, this task remains highly challenging. It requires accurately capturing a writer's unique and variable style—including features like stroke width, curvature, and slant, plus subtle nuances like ink density—from a single image. Despite recent advancements in generative models, current methods still struggle to extract these \textbf{core stylistic attributes} from unseen writers and produce \textbf{high-quality} results. Consequently, the generated image is often stylistically incomplete; for example, a model might replicate a writer's slant but fail to capture the variations in their ink pressure. This leads to the central question: \textit{How can we effectively capture the complex stylistic characteristics from a single handwriting sample to generate diverse, realistic, and high-quality text?}

Previous approaches for handwriting synthesis have largely relied on Generative Adversarial Networks (GANs)~\cite{fogel2020scrabblegan,ganwriting_eccv_2020,HiGan_Wang_2021,HiGANplus,VATr_2023_CVPR,HWT_2021_ICCV}. While capable of generating diverse samples, GANs often struggle to produce realistic images, especially for complex writing styles, and are known for unstable training.

More recently, Diffusion Models (DMs)~\cite{ddpm_nips_2020,rombach2022high,dhariwal2021diffusion} have emerged as a powerful alternative, demonstrating superior performance over GANs in both quality and diversity~\cite{CTIGDM_2023_CVPR,nikolaidou2023wordstylist,dai2025one,nikolaidou2024diffusionpen}. However, a key limitation remains: the inability to comprehensively model style information. For instance, One-DM~\cite{dai2025one}, a current state-of-the-art (SOTA) one-shot DMs method, uses a high-frequency style encoder that overlooks important features like stroke density and color, while other approaches~\cite{HWT_2021_ICCV,VATr_2023_CVPR} employ computationally intensive Transformer-based encoders~\cite{vaswani2017attention}. To improve quality, many methods~\cite{VATr_2023_CVPR,HWT_2021_ICCV,nikolaidou2024diffusionpen} resort to few-shot generation (using many reference images), a setting that is less practical than the one-shot approach.

In this work, we focus on improving one-shot handwritten text generation (HTG) performance by introducing a novel approach called Patch \textbf{Con}trastive Enhancement and \textbf{St}yle-\textbf{A}ware Qua\textbf{nt}ization via Denoising Diffusion (\textbf{CONSTANT}). Our method harnesses vector quantization (VQ), a simple yet effective approach, to represent complex handwriting style information as discrete visual tokens, where each token corresponds to a fundamental style concept within an embedding space, mirroring how humans might intuitively categorize handwriting traits (see Fig.~\ref{fig:mainidea}a). Unlike methods like One-DM~\cite{dai2025one} that rely on fixed high-frequency filters, our learned \say{style concepts} are more robust at capturing the core characteristics from a single reference image while effectively filtering out incidental noise (see Fig.~\ref{fig:mainidea}b). We also incorporate a style contrastive enhancement objective, $L_{SCE}$, to refine the latent space by promoting similarity among style features from the same writer while distinguishing them from those of different writers. To boost synthesis quality and avoid issues like blurriness or inconsistencies between target and generated outputs during training, we propose a Latent Patch Contrastive Enhancement ($L_{LatentPCE}$), an auxiliary objective that goes beyond the standard denoising loss used by One-DM~\cite{dai2025one} by using contrastive learning to align the latent features of generated and target images, sharpening local details for better results by maximizing mutual information between the locations of corresponding target and generated patches. Unlike multi-stage training in works like~\cite{HiGANplus,dai2025one}, our entire process operates in a single, end-to-end stage. To assess our method on Vietnamese, we created a new HTG dataset, \textbf{ViHTGen}, specifically for Vietnamese. We tested our approach across multiple datasets in various languages, including English, Chinese, and our proposed ViHTGen, to showcase its effectiveness. In summary, our contributions are: 1) We present a Style-Aware Quantization (SAQ) module that treats each style concept as a discrete visual token for extracting handwriting styles, enabling clearer separation of style representations, thus helping reduce the loss of extracted style features when it comes to complex style reference images. Paired with a style contrastive objective ($L_{SCE}$), it strengthens distinctions between different styles. 2) We propose a new contrastive learning objective $L_{LatentPCE}$ that leverages patch-level latent features in the DMs to refine local details and maintain consistency in generated images. 3) We performed a thorough evaluation on multiple language datasets, including our new ViHTGen dataset, confirming our method's state-of-the-art performance in terms of visual quality, style quality, and readability.
\section{Related Works}

\subsection{Handwritten Text Generation}
Early methods~\cite{fogel2020scrabblegan,ganwriting_eccv_2020,SmartPatch,HiGan_Wang_2021,HiGANplus,HWT_2021_ICCV,VATr_2023_CVPR,jokergan_2021} primarily relied on GANs due to their capability to model complex data distributions inherent in handwriting. In GANs, one-shot generation methods like ScrabbleGAN~\cite{fogel2020scrabblegan} utilize random noise as style input to synthesize handwritten images, thus unable to control the writers' style. HiGAN~\cite{HiGan_Wang_2021} proposes a CNN-based style encoder to extract style features combined with a text recognition loss to improve quality. This line of work was further refined by HiGAN+~\cite{HiGANplus}, which incorporated a contextual loss and a PatchGAN-based~\cite{demir2018patch} discriminator, and SmartPatch~\cite{SmartPatch}, which employed a similar lightweight patch discriminator.

More recently, the field has shifted towards DMs for their ability to generate high-quality, diverse images. Models like WordStylist~\cite{nikolaidou2023wordstylist} and CTIG-DM~\cite{CTIGDM_2023_CVPR} demonstrated the potential of DMs but were limited by their reliance on predefined writer IDs, preventing generalization to unseen styles. Similarly, font-focused DMs like Diff-Font~\cite{he2024diff} and FontDiffuser~\cite{yang2024fontdiffuser} apply diffusion with contrastive elements for one-shot style adaptation, yet target structured glyphs rather than variable handwriting. The most relevant one-shot diffusion model, One-DM~\cite{dai2025one}, addresses this by using an enhanced style encoder; however, its reliance on a predefined Laplacian kernel to extract high-frequency information can be brittle. Our approach overcomes this limitation with two key contributions. First, our SAQ module learns an adaptive, discrete vocabulary of \say{style concepts}, making it more robust to the diverse and noisy styles found in real-world images. Second, we introduce $L_{LatentPCE}$ objective to specifically improve the local detail and realism of the generated images.

 Alternative techniques employ a few-shot strategy to enhance style imitation. GANWriting~\cite{ganwriting_eccv_2020} employs 15 reference images to extract style features. In contrast, HWT~\cite{HWT_2021_ICCV} adopts a Transformer architecture instead of CNN, implementing an encoder-decoder Transformer framework to generate stylized text. VATr~\cite{VATr_2023_CVPR} and VATr++~\cite{vanherle2024vatr++} refine this by treating characters as continuous query vectors in a Transformer decoder. DiffusionPen~\cite{nikolaidou2024diffusionpen} utilizes DMs combined with a metric learning objective to improve generated quality but still relies on 5-shot reference images. Departing from these methods, our work concentrates on the one-shot learning paradigm, which presents significant challenges while offering greater practical applicability.

% \subsection{Conditional Diffusion Models}
% DMs are generative models trained to learn a specific distribution from data by denoising variables sampled from a
% Gaussian distribution in T steps. To reduce the considerable
% computational cost, LDMs perform this denoising process
% within a perceptually compressed latent space encoded by a
% pre-trained VAE.
% Conditional diffusion models represent a significant advance in generative modeling, often surpassing GANs in various tasks. Starting with DDPM~\cite{ddpm_nips_2020}, which uses iterative denoising to generate samples, numerous improvements have enhanced quality and control~\cite{ddim_iclr_2021,dhariwal2021diffusion,rombach2022high}. Techniques like classifier-free guidance (CFG)~\cite{ho2022classifier} and multimodal conditioning as seen in GLIDE~\cite{nichol2022glide} have further boosted performance, especially in text-to-image generation. Nevertheless, although these models demonstrate remarkable proficiency in generating diverse general images, their direct application to handwriting synthesis is often less successful. Handwriting demands imitation of fine-grained stylistic details and structural consistency specific to an individual, aspects that pose challenges for models primarily designed for broader visual feature generation.

\subsection{Disentangled Representation Learning (DRL)}
DRL~\cite{wang2024disentangled} focuses on creating models that identify and separate the underlying factors within data into distinct, semantically meaningful representations. Based on that concept, some works including~\cite{friede2023learning,hsu2023disentanglement} further explore specific inductive biases and model architectures to enhance DRL by using discrete representations where each discrete value can be assigned a consistent meaning. This concept has been applied to tasks like text image super-resolution. For example, \cite{li2023learning} employs a codebook for Chinese characters, and \cite{chen2021exploring} trained a VAE where representations of similar letters are closer. Different from those works, our approach aims to leverage discrete representations through VQ to model complex style concepts in handwriting, which is more difficult than the global character structure of~\cite{li2023learning,chen2021exploring}.
\section{Methodology}
\begin{figure*}[tbh]
  \centering
    \includegraphics[width=0.8\textwidth]{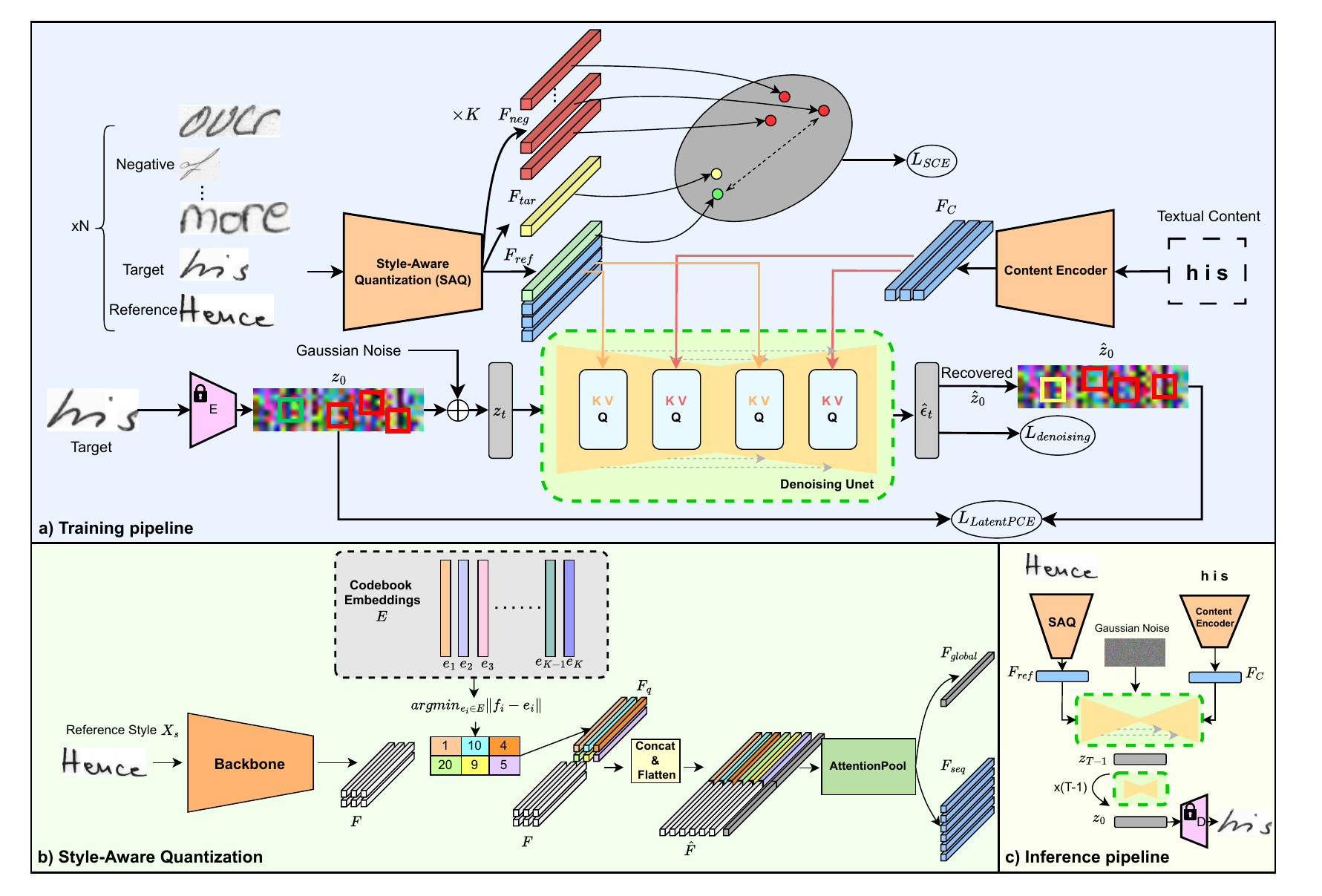}
    \caption{\textbf{Overall architecture of our method}. a) The training pipeline, the model is optimized using objectives: $L_{denoising}$, $L_{SCE}$ and $L_{LatentPCE}$ and $L_{SAQ}$, b) Architecture of our SAQ module, with Inception-V3 backbone, codebook embedding and a \textit{Attention Pool} module, c) Sampling pipeline of our method.}
    \label{fig:main_figure}
    \vspace{-0.15in}
\end{figure*}

\subsection{Problem Definition}
The purpose of a one-shot handwriting generation model is to generate any handwritten text images conditioned on a reference style image and textual content information. Mathematically, given a reference image $X_s$ from a writer $W_s$, and a word input $C=\{c_i|c_i\in \mathcal{A}\}_{i = 1}^{L}$, where $\mathcal{A}$ is the alphabet, $L$ is the length of word input, our goal is to generate new image $X_t$, subscript \say{t} stand for \say{target}, that represent textual content $C$ that does not depend on any predefined corpus with the same handwritten style as $X_s$.
% Our \textbf{CONSTANT} model aims to enhance style extraction by leveraging VQ for style extraction through SAQ module with support from $L_{SCE}$ object and generation high-quality and robustness with out-of-vocabulary (OOV) words by a novel $L_{LatentPCE}$ as an auxiliary objective beside denoising loss of LDM.
% \vspace{-0.1in}
\subsection{Overall Architecture}
% In the training phase, our proposed  $L_{LatentPCE}$ improves the local consistency between the target image and generated image by maximizing mutual spatial information at the specific patch-level location of reconstructed features and corresponding target features. 
 Our method utilizes Latent Diffusion Models (LDMs)~\cite{rombach2022high}, which are more computationally efficient and use less memory than traditional pixel-level models~\cite{ddpm_nips_2020,dhariwal2021diffusion} that process high-resolution images directly. As shown in Fig.~\ref{fig:main_figure}a, our approach conditions LDMs on style and textual features obtained from a style extractor and content encoder. To be able to synthesize handwriting images from an unseen style reference, it is important that the style extraction module has knowledge of the diverse style concepts in the handwriting domain and represents each style concept distinctively. We utilize VQ technique with a pre-defined codebook embedding to discretize the extracted style features into different discrete visual tokens, followed by many works in discrete representation learning~\cite{van2017neural,mao2021discrete,peng2022beit}, helping our model to learn the necessary information and ignoring noise information more effectively. The codebook embedding is optimized using an $L_{SAQ}$ objective.  \\
Furthermore, to better encourage our model to learn mutual information between different samples from the same writer and discourage irrelevant features, we leverage $L_{SCE}$, which receives a global style representation after combining different style concepts from SAQ module to discover a more discriminative style embedding space as shown in Fig.~\ref{fig:main_figure}b. Besides the denoising objective $L_{denoising}$ of LDMs\footnote{Further details on the latent diffusion models will be provided in the Appendix}, our $L_{LatentPCE}$ helps to preserve the visual quality and details of generated images inspired by~\cite{andonian2021contrastive}. For the textual content encoding, we utilize a 3-layer Transformer~\cite{vaswani2017attention} encoder, our content encoder represents each input character as an embedding vector independently, specifically, given the input word $C=\{c_i\}_{i=1}^L$, our encoder extract an embedding vector $F_c \in \mathbb{R}^{L\times d}$, where d is the channel dimension, which will allow our model more robust and be able to generate OOV words more easily. Both style and content features are then input as the context information to the cross-attention module inside the DM. Overall, our proposed method is jointly optimized in an end-to-end manner by the combination of the following objectives, as illustrated in 
Fig.~\ref{fig:main_figure}a:
\begin{equation}
    L = L_{denoising} + \alpha \times (L_{LatentPCE} + L_{SCE} + L_{SAQ})
\end{equation}
Where $\alpha$ is the weight of all supported objectives, empirically, we choose $\alpha = 0.1$.

\subsection{Style-Aware Quantization}
\label{sec:saq}
As discussed in the above section, we propose a SAQ module to extract compact and discriminative style representations from handwriting images. Our core motivation is to move beyond monolithic, continuous style vectors that can overfit to noise in a \textbf{single reference image}. To achieve this, we propose to take advantage of discrete representation ability of VQ~\cite{van2017neural} to capture different style concepts as discrete embedding tokens inside a codebook embedding that covers a wide range of style variations in handwriting. Our proposed SAQ employs a pretrained InceptionV3~\cite{szegedy2016rethinking} backbone, leveraging its multi-scale feature extraction capabilities from ImageNet~\cite{deng2009imagenet}, widely utilized in style transfer~\cite{ghiasi2017exploring}. Given a reference style image $X_s \in \mathbb{R}^{H\times W \times 3}$, the backbone first creates a feature map $F \in \mathbb{R}^{h \times w \times d}$, where d is the feature dimension. A codebook embedding $E=\{e_i\}_{i=1}^K \in \mathbb{R}^{K\times D}$, where K is the length of the codebook, which stands for a collection of different style concepts and $D$ is the codebook dimension, is used to map each feature in the feature maps to a specific discrete code a.k.a, visual token. A larger value of $K$ could model finer-grained traits (e.g., subtle ligature variations), but risks overfitting to noise or increasing memory demands, while a smaller value might oversimplify styles, missing nuanced writer-specific characteristics. Let $F_q=\{f_q^i\} \in \mathbb{R}^{h \times w \times d}$ is the quantized feature maps, each vector $f_q^i$, where $i \in [0, h\times w ]$ of $F_q$ is calculated by looking up the nearest neighbor of $f_i$ in the codebook $E$ as  
\begin{equation}
    f_q^i = argmin_{e_i \in E} \|f_i - e_i \|_2
\end{equation}
Similar to other previous methods~\cite{van2017neural,mao2021discrete,peng2022beit}, we optimize codebook embedding $E$ using VQ-loss $L_{vq}$ and commitment loss $L_{cmt}$ to ensure the codebook learns meaningful clusters of style features.
\begin{equation}
    L_{SAQ} = L_{vq} + \beta\times L_{cmt}
\end{equation}
Where loss terms $L_{vq} = \left\|\operatorname{sg}\left(F\right) - E\right\|_2^2$, $L_{cmt} = \left\|F - \operatorname{sg}\left(E\right)\right\|_2^2$, where $\operatorname{sg}$ stands for the stop gradient operator, and $\beta$ is the weight of commitment loss, we choose $\beta = 0.25$ similar to~\cite{van2017neural}.
Nevertheless, we noticed that directly using quantized feature map $F_q$ and completely ignoring all information from $F$ will lead to the ignoring of some useful local details information, as the quantized feature maps only capture general style concepts that should not be missing when extracting from the reference images, but the continuous features are also useful to be able to adapt to each specific unseen writer style. Therefore, we concatenate the continuous feature maps $F$ with the quantized version $F_q$ and flatten the result into $\hat F \in \mathbb{R}^{T\times 2d}$, where $T = h\times w$ and apply the \textit{Attention Pool} module inspired by~\cite{radford2021learning} to fuse the information from both feature maps. This module performs a self-attention operation on $\hat F$ producing two types of features: 1) $F_{global} \in \mathbb{R}^{d}$ as the global representation of $\hat F$, used with $L_{SCE}$ to enhance writer discrimination; 2) $F_{seq} \in \mathbb{R}^{T\times d}$ as the sequence of fused style features after refinement serving as context for the DM. This hybrid approach balances the abstraction of discrete tokens with the richness of continuous features, making it robust to the variability in handwritten images. The overall architecture of our SAQ is illustrated in Fig.~\ref{fig:main_figure}b.
% \vspace{-0.1in}
\subsection{Style Contrastive Enhancement}
\label{sec:sce}
% Contrastive learning is a powerful self-supervised learning technique that aims to learn meaningful representations from data~\cite{le2020contrastive,tian2020contrastive,park2020contrastive,andonian2021contrastive}, by contrasting positive and negative pairs of data samples, the model is encouraged to learn representations that capture the underlying structure and similarities within the data. With such powerful characteristics, we propose to apply a contrastive objective to style feature learning process to encourage style features of the same writers to be closer in embedding space from those of other writers. 
Given a training batch $B$ with $N$ samples, including $N/2$ pairs of (reference, target) images from different writers. Let target (anchor) style feature as $F_{tar} \in \mathbb{R}^{d}$ with index j and reference style feature as $F_{ref} \in \mathbb{R}^{d}$ with index p, for each (reference, target). We have $K=N-2$ negative samples, we call them $D = B \setminus \{j,p\}$, where $F_d \in \mathbb{R}^{Kxd}$ is the style feature of negative samples. Both $F_{tar}$, $F_{ref}$, and $F_d$ are the global style feature $F_{global}$ returned by SAQ module. Our style contrastive enhancement ($L_{SCE}$) loss is written as follows:

\begin{equation}
\resizebox{.9\linewidth}{!}{$
            \displaystyle
    \ell_{SCE}\left(F_{tar}, F_{ref}, F_d\right)=\frac{-1}{N} \sum_{t \in B} \log \frac{\exp \left(\operatorname{sim}\left(F_{tar}, F_{ref}\right) / \tau\right)}{\exp \left(\operatorname{sim}\left(F_{tar}, F_{ref}\right) / \tau\right) + \sum_{d \in D(t)} \exp \left(\operatorname{sim}\left(F_{tar}, F_d\right) / \tau\right)}
    $}
\end{equation}

Where $\operatorname{sim}\left((F_{tar}, F_{ref}) / \tau\right)$ is the cosine similarity between reference and target images and $\tau$ a temperature hyper-parameter.
More generally, we can swap the role of the reference and target as different views as discussed in \cite{tian2020contrastive}, where the reference is Anchor and the target is Positive sample. Therefore the overall $L_{SCE}$ is as follows:
\begin{equation}
\resizebox{.9\linewidth}{!}{$
    L_{SCE} = \frac{1}{2}\ell_{SCE}\left(F_{tar}, F_{ref}, \operatorname{sg}\left(F_d\right)\right) + \frac{1}{2}\ell_{SCE}\left(F_{ref}, F_{tar}, \operatorname{sg}\left(F_d\right)\right)
    $}
\end{equation}
Here $\operatorname{sg}$ indicates a stop gradient operation, which prevented gradients from going through negatives as proposed from~\cite{chen2021exploring}. \\

\subsection{Latent Patch Contrastive Enhancement}
\label{sec:laentpce}
In HTG, producing realistic and fine-grained detailed outputs is crucial. Traditional regression losses, such as standard denoising loss in LDMs, help generate globally coherent results but often struggle to capture fine details, sometimes resulting in blurry or oversmoothed outputs. Our proposed $L_{LatentPCE}$ is a complementary loss to the denoising loss, ensuring local consistency and enhancing perceptual realism. Unlike HiGAN+~\cite{HiGANplus}, which uses pixel-level patching to enhance real/fake sample classification, our $L_{LatentPCE}$ operates in the latent space to directly optimize the locality information of generated images.

\vspace{-0.15in}
 \begin{figure}[H]
  \centering
    \includegraphics[width=0.9\linewidth]{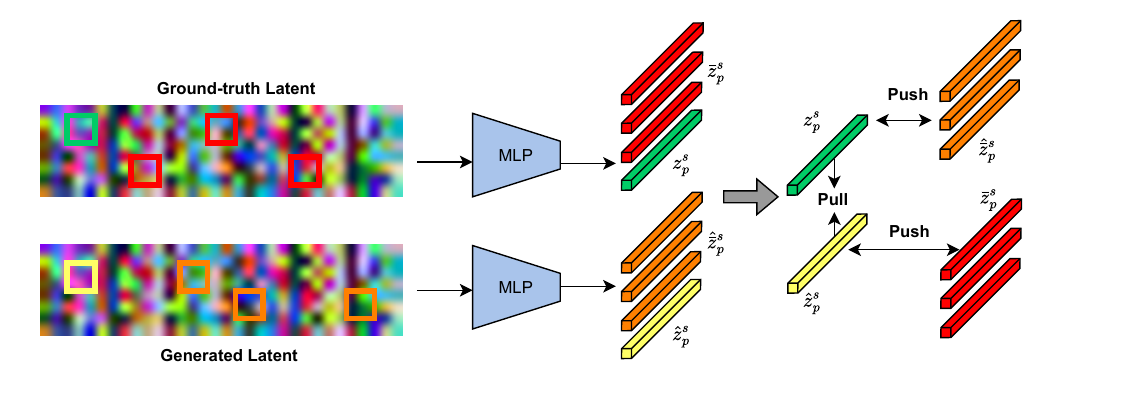}
    \caption{Visualization of our $L_{LatentPCE}$ objective in bidirectional format including two sub contrastive loss. First loss receives (\textcolor{green}{anchor}, \textcolor{yellow}{positive}, \textcolor{orange}{negatives}), second loss receives (\textcolor{yellow}{anchor}, \textcolor{green}{positive}, \textcolor{red}{negative}).}
    \label{fig:pceloss}
    \vspace{-0.1in}
\end{figure}

 Given a ground-truth and a generated latent feature, we extract $y_p, \hat y_p \in \mathbb{R}^{h_{patch}\times w_{patch}\times d_{patch}}$ from the same spatial location (e.g. \textcolor{green}{green} \& \textcolor{yellow}{yellow} in Fig.~\ref{fig:pceloss}). Similarly, $\bar y_{p}, \hat{\bar y}_{p}$ correspond to other spatial patches in the ground-truth and generated features (e.g. \textcolor{red}{red} \& \textcolor{orange}{orange} in Fig.~\ref{fig:pceloss}). These patches are projected into feature embeddings $z_p, \hat z_p \in \mathbb{R}^D$, $\bar{z}_p$, $\hat{\bar{z}}{}_p \in \mathbb{R}^{(H-1)\times D}$ using a shallow MLP network. Where $D$ is channel dimension, $H$ is the total number of patches. The objective is to ensure that patches from the same spatial location are closer in the embedding space, while patches from different locations are pushed farther apart. Additionally, we extract patches at multiple scales by varying patch sizes to capture local information across different spatial resolutions. In summary, our $\ell_{LatentPCE}$ at certain scale level $s$ is written as follows:
 
\begin{equation}
\vspace{-0.1in}
\resizebox{.9\linewidth}{!}{$
    \ell_{LatentPCE}(z_p^s, \hat z_p^s, \hat{\bar{z}} {}_{p}^{s}) = \\ -\frac{1}{H}\sum_{p=1}^{H}\log \frac{\exp \left(sim\left(z_p^s, \hat z_p^s\right) / \tau\right)}{\exp \left(sim\left(z_p^s, \hat z_p^s\right) / \tau\right)+\sum_{n=1}^{H_s-1} \exp \left(sim\left(z_p^s, \hat {\bar{z}}{}_p^s\right) / \tau\right)}   
    $}
\end{equation}
 
Similar to $L_{SCE}$, we can write $L_{LatentPCE}$ in a bidirectional format by swapping roles of the anchor and positive samples. Therefore, $L_{LatentPCE}$ objective at multiple scales can be written as follows:

\begin{equation}
\vspace{-0.1in}
\resizebox{.9\linewidth}{!}{$
   L_{LatentPCE} = \frac{1}{2S}\sum_{s=1}^{S}\ell_{LatentPCE}(z_p^s, \hat z_p^s, \operatorname{sg}(\hat{\bar{z}}{}_p^s)) + \frac{1}{2S}\ell_{LatentPCE}(\hat z_p^s, z_p^s, \operatorname{sg}({\bar{z}_p^s}))
   $}
\end{equation}
Here $\operatorname{sg}$ indicates a stop gradient operation similar to $L_{SCE}$, and $S$ is the number of scale levels.

\section{Experiments}
% In this section, we evaluate our method on several benchmark datasets and compare its performance with current state-of-the-art (SOTA) methods. We conduct extensive analysis to demonstrate the effectiveness of our key components. An in-depth analysis is provided to show the ability to extract useful style information and generate high-quality handwritten text. We also evaluate HTR tasks to show the potential of our generated samples on other downstream tasks in practice.

\subsection{Experiment Settings}

\subsubsection{Datasets}
We evaluate our method on three datasets, including the standard \textbf{IAM} dataset~\cite{marti2002iam}, a real-world complex dataset as \textbf{IMGUR5k}~\cite{krishnan2021textstylebrush}, and a new \textbf{IIIT-English-Word} dataset~\cite{mondal2024bridging}. \textbf{IAM} dataset consists of 62,857 English word images from 500 unique writers. Similar to the setting in GANWriting~\cite{kang2020ganwriting}, we use data from 339 writers for the training phase and the remaining 161 writers for evaluation. \textbf{IMGUR5K} dataset is a real-world multi-source domain dataset provided with a large variety of handwritten styles, containing 135,375 words from 5305 writers. \textbf{IIIT-English-Word} dataset is a new dataset containing a large and diverse set of handwritten words with more than 700000 samples from 1,215 writers. \textit{Results on \textbf{IIIT-English-Word} are presented in the Appendix.}
% More details about these benchmark datasets are provided in Appendix~\ref{sec:dataset_supp}.

\subsubsection{Evaluation Metrics}
To evaluate the visual quality of our proposed method and compare it with other SOTA methods, we use the Fréchet Inception Distance (FID)~\cite{heusel2017gans}. For style imitation quality, we use two metrics: Handwriting Distance (HWD)~\cite{pippi2023hwd} to measure the perceptual distance of subtle geometric features between the generated and real images using Euclidean distance, and writer classification accuracy a.k.a $Acc_{Wid}$, obtained by testing a classifier trained on real images with our generated samples. For readability, we follow the setup from~\cite{nikolaidou2024rethinking} to assess handwritten text recognition performance via Word Error Rate (WER). \textit{Finally, these quantitative results are supplemented by a user study on perceptual quality and diversity, with full details provided in the Appendix.}

\subsubsection{Implementation Details}
For all of our experiments, we train our model in only one stage using PyTorch 1.12.1 for 800000 iterations using batch size of 64 for IAM dataset and 128 for IMGUR5K and IIIT-English-Word dataset. All images are resized to a fixed height of 64 and keep the aspect ratio. We hypothesize that datasets with greater complexity—such as noisier backgrounds and foreground variations—require a larger codebook size K to capture more diverse style concepts. Empirically, we thus set $K=1024$ for IAM and $K=2048$ for IMGUR5K/IIIT-English-Word, with detailed ablation studies validating this across datasets provided in the Appendix. Training process is conducted with AdamW optimizer~\cite{loshchilov2017decoupled}, with $\beta_1=0.9$, $\beta_2=0.999$, and an initial learning rate of $10^{-4}$ on a single Nvidia V100-32GB GPU. During training phase, we follow the idea of classifier-free guidance (CFG)~\cite{ho2022classifier} to randomly drop conditional features with probability $p=0.2$. At inference phase, as shown in Fig.~\ref{fig:main_figure}c, we adopt CFG strategy with a guidance scale of 7.5 using denoising diffusion implicit model (DDIM)~\cite{ddim_iclr_2021} with 50 sampling steps~\footnote{More details about our implementation and architecture are provided in the Appendix.}.

\subsubsection{Compared with Other SOTA}
We compare our method against several SOTA handwritten text generation (HTG) methods. GAN-based comparisons include HiGAN~\cite{HiGan_Wang_2021}, HiGAN+~\cite{HiGANplus}, HWT~\cite{HWT_2021_ICCV}, and VATr~\cite{VATr_2023_CVPR}. For diffusion model (DM)-based methods, we evaluate against the current SOTA One-DM~\cite{dai2025one} and DiffusionPen~\cite{nikolaidou2024diffusionpen}. Among these, HiGAN, HiGAN+, and One-DM are one-shot HTG methods, while HWT, VATr, and DiffusionPen are few-shot~\footnote{We utilize pretrained weights for IAM where available; otherwise, we adapt provided source code to train models from scratch on other datasets.}.

\subsection{Compared with SOTA Methods}
\subsubsection{Quantitative Comparison}
Tab.~\ref{tab:iamcompare} presents the quantitative results on the IAM test set. Our method achieves state-of-the-art performance across both one-shot and few-shot approaches, with the best HWD of 0.74 and FID of 10.20. This outperforms the second-best HiGAN+ (0.89 HWD, 13.90 FID). As a one-shot DM-based method, ours also surpasses One-DM in both HWD and FID scores. For readability, we achieve a SOTA WER of 0.22, marginally better than DiffusionPen's 0.23 and significantly outperforming One-DM's 0.36. These results demonstrate the effectiveness of our $L_{LatentPCE}$ (Sec.~\ref{sec:laentpce}). Furthermore, the proposed SAQ (Sec.~\ref{sec:saq}) module enhances style quality, evidenced by a 69.43\% writer classification accuracy.  \\
Furthermore, we also evaluate our approach in four different scenarios, including IV-S (In-vocab, Seen style), OOV-S (Out-of-vocab, Seen style), IV-U (In-vocab, Unseen style), OOV-U (Out-of-vocab, Unseen style), where OOV-U is considered the most difficult case, similar to settings from \cite{kang2020ganwriting,HWT_2021_ICCV}. As shown in Tab.~\ref{tab:styleHTG}, our method demonstrates strong performance in terms of both FID and HWD scores. While DiffusionPen excels among GAN-based methods, particularly in HWD, highlighting the strength of diffusion models for style adaptation, our approach achieves state-of-the-art results in seen style cases (IV-S and OOV-S). We also establish a significant lead over DiffusionPen in both HWD and FID, despite using only a single reference image. For unseen cases, our method outperforms the second-best (DiffusionPen) on HWD for both IV-U and OOV-U, and exceeds One-DM on FID for OOV-U by more than 3\%. These results clearly demonstrate the effectiveness of our method, especially the SAQ module (Sec.~\ref{sec:saq}), in generalizing to unseen writing styles.\\
Beyond the IAM dataset, we also evaluated our method on the more complex IMGUR5K dataset. As shown in Tab.~\ref{tab:imgur5kcompare}, our approach achieves state-of-the-art results on the IMGUR5K test set, with an HWD score of 0.99, outperforming One-DM's 1.22. This demonstrates our method's superior capability in replicating complex handwritten styles using the SAQ module, compared to One-DM's high-frequency encoder.

\begin{table}[t]
    \vspace{-0.2in}
    \begin{center}
    \scalebox{0.65}{
    \large
    \begin{threeparttable} 
        \begin{tabular}{c||c|c|c|c|c}
        \toprule
        Method      & Reference &  HWD $\downarrow$       & FID  $\downarrow$ & WER $\downarrow$ & $Acc_{Wid}$ $\uparrow$     \\
        \midrule
        HWT~\cite{HWT_2021_ICCV}  & \multirow{3}{*}{Few-shot}       &  1.23   & 19.82 &  0.62 & 9.08 \\
        VATr~\cite{VATr_2023_CVPR}   &     &  1.13   & 16.30  & 0.51 & 49.81\\
        DiffusionPen~\cite{nikolaidou2024diffusionpen} &  & 1.04 & 18.94 & \underline{0.23} & 38.86 \\
        \midrule
        HiGAN+~\cite{HiGANplus}  &  \multirow{4}{*}{One-shot}  & \underline{0.89}   & \underline{13.90} & 0.56 & \underline{55.20}  \\
        HiGAN~\cite{HiGan_Wang_2021}  &     &  1.55   & 27.13  & 0.55 & 29.79 \\
        % WordStylist \cite{nikolaidou2023wordstylist} &  0.89  & 17.41  \\
        One-DM \cite{dai2025one}  &    & 1.05 & 15.97 & 0.36 & 4.5\\
        % \midrule
        \textbf{Ours}    &  & \textbf{0.74}   & \textbf{10.20}  & \textbf{0.22} & \textbf{69.43} \\
        \bottomrule
        \end{tabular}
    \end{threeparttable}    
    }
\caption{Comparison between SOTA methods on IAM test dataset. \textbf{Bold} values indicate best and \underline{Underline} values indicate second-best.} \label{tab:iamcompare}
    \vspace{-0.3in}
    \end{center}
\end{table}
 
\begin{table}[thb] 
\centering
    \vspace{-0.1in}
   \scalebox{0.5}{
    \large
    \begin{threeparttable} 
        \begin{tabular}{c||*{2}{c}|*{2}{c}|*{2}{c}|*{2}{c}}
        \toprule
                        & \multicolumn{2}{c}{IV-S} & \multicolumn{2}{c}{OOV-S} & \multicolumn{2}{c}{IV-U} & \multicolumn{2}{c}{OOV-U} \\
                     Method   & HWD $\downarrow$       & FID $\downarrow$          & HWD $\downarrow$       & FID $\downarrow$         & HWD $\downarrow$       & FID $\downarrow$         & HWD $\downarrow$         & FID $\downarrow$         \\
                       \midrule
            HWT~\cite{HWT_2021_ICCV}   & 2.30   & 135.51      & 2.30    &   146.16   & 2.35    &  138.39   & 2.36    &   148.75  \\
            VATr~\cite{VATr_2023_CVPR}  & 2.50   &   132.87    & 2.51     &    140.56    & 2.58     &  137.34   & 2.60      &    144.02    \\
            DiffusionPen~\cite{nikolaidou2024diffusionpen} & \underline{1.14} & \underline{91.20} & \underline{1.16} & \underline{97.65} & \underline{1.52} & 112.87 & \underline{1.65} & 122.52 \\
            \midrule
            HiGAN~\cite{HiGan_Wang_2021}  &  2.29 & 118.70      & 2.31  &   128.49  & 2.36      &  119.56   &  2.37    & 128.60  \\
            HiGAN+~\cite{HiGANplus} &  1.76 & 117.76       & 1.79    & 122.56    & 1.78    & \underline{117.63}     & 1.81   & 124.38   \\
            % WordStylist~\cite{nikolaidou2023wordstylist}  & \underline{1.62}    & \underline{97.46}      & \underline{1.63}     & \underline{102.99}      & 2.17     & \underline{114.44}   & 2.22      & \underline{119.43}    \\
            % \midrule
            One-DM~\cite{dai2025one} &     1.95 & 104.04      & 1.99  & 107.81    & 1.94   & 117.74    & 1.99     & \underline{121.94}    \\
            \textbf{Ours (CONSTANT)}  &     \textbf{0.96}  & \textbf{89.88}  & \textbf{0.94}    & \textbf{96.13} & \textbf{1.61}      & \textbf{112.03}   &  \textbf{1.63} &  \textbf{118.10}\\
            \bottomrule
        \end{tabular}
        % \vspace{-0.1in}
        \end{threeparttable}    
    }
    \caption{Evaluation in terms of FID and HWD scores on four scenarios: IV-S (In-vocab, Seen style), OOV-S (Out-of-vocab, Seen style), IV-U (In-vocab, Unseen style), OOV-U (Out-of-vocab, Unseen style). \textbf{Bold} values indicate best and \underline{Underline} values indicate second-best.}
    \label{tab:styleHTG}
\end{table}

\begin{table}[tbhp]
    \vspace{-0.2in}
    \centering
    \small
\begin{minipage}{0.48\linewidth}
    \begin{center}
    \scalebox{0.65}{
        \begin{tabular}{c||c|c}
        \toprule
        Method      & HWD  $\downarrow$   & FID  $\downarrow$      \\
        \midrule
        HiGAN+ \cite{HiGANplus}      & 1.35 & 20.04   \\
        HiGAN \cite{HiGan_Wang_2021}       &   1.55      & \underline{17.58}  \\
        % WordStylist \cite{nikolaidou2023wordstylist} & 1.16  & 24.25    \\
        One-DM \cite{dai2025one}     & \underline{1.22}  & 18.94     \\
        \midrule
        \textbf{Ours}    & \textbf{0.99}  & \textbf{11.48} \\
        \bottomrule
        \end{tabular}
    }
    % \vspace{-0.1in}
    \end{center}
    \caption{Quantitative results on IMGUR5K test set.} 
    \label{tab:imgur5kcompare}
\end{minipage}
% \hspace{0.05in}
\hfill
\begin{minipage}{0.48\linewidth}
    \centering
    
        % \vspace{4pt} % Add a little vertical space between caption and table
    \scalebox{0.65}{
        \begin{tabular}{c||c}
        \toprule
        Method &  FID $\downarrow$ \\
        \midrule
        Baseline & 16.73 \\
        Baseline + $L_{cosine}$ & 21.06 \\
        Baseline + $LPIPS$ & 14.05 \\
        Baseline + $L_{PatchL2}$ & 15.10 \\
        \textbf{Baseline + $L_{LatentPCE}$}  & \textbf{14.01} \\

        \bottomrule
    \end{tabular}
    }
    \caption{Quatitative comparisons between $L_{LatentPCE}$ and other auxiliary objectives in terms of FID.}
    \label{tab:pcevscosinetab}
\end{minipage}
\end{table}

\vspace{-0.1in}
\subsubsection{Qualitative Comparison}
\begin{figure*}[tbh]
    \centering 
\includegraphics[width=0.85\textwidth]{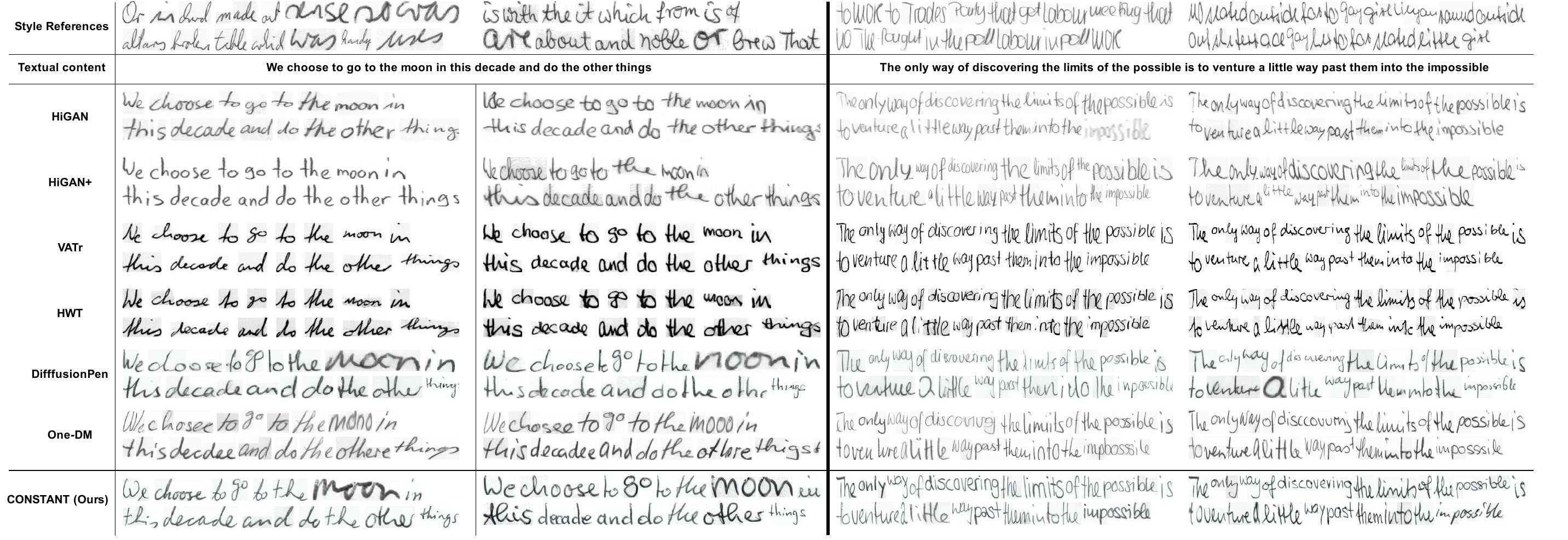}
    \caption{Comparison between ours with other methods using random style from IAM test set and two different textual contents.}
    \label{fig:wordvis}
    \vspace{-0.15in}
\end{figure*}

\begin{figure}[tbh]
    \centering 
\includegraphics[width=0.8\linewidth]{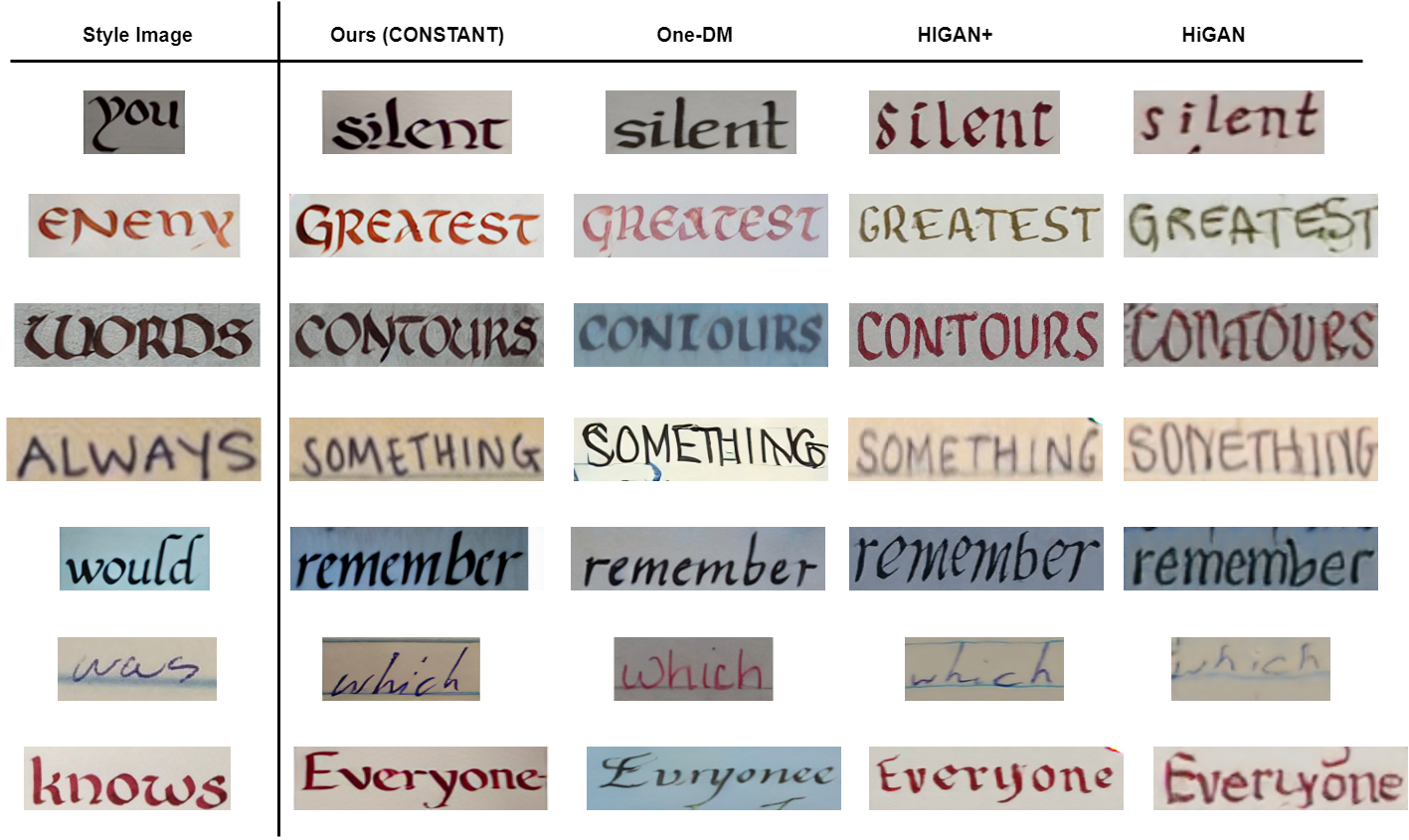}
    \caption{Visualization of handwritten text generation on IMGUR5K test set.}
    \label{fig:imgur5kviz}
    % \vspace{-0.1in}
\end{figure}

Fig.~\ref{fig:wordvis} presents a qualitative comparison, showcasing our method's ability to generate new content with a style (slant, stroke width, and background color) closely matching reference images. In contrast, HiGAN and HiGAN+ fail to capture the slant variation. Although HWT and VATr utilize 15 times more reference images than our method, they struggle with accurate character color and shape adaptation, and inconsistencies often arise from high intra-writer variability in their references. While One-DM and DiffusionPen extract style better than other GAN-based methods, they fall short in generating correct textual content and sufficient local detail. Fig.~\ref{fig:imgur5kviz} visually demonstrates the superiority in mimicking style, including ink color and character shapes, on IMGUR5K dataset. We can also see that One-DM with only high-frequency information can generate reasonable results, still can not learn the ink color information of the style image effectively.

% \vspace{-0.12in}

\subsection{Ablation Studies}
\subsubsection{Effectiveness of Proposed Components}
\label{sec:abla}
To validate our contributions, we perform a cumulative ablation study, with results shown in Tab.~\ref{tab:componentablation}. We start with a \say{Base} model, a standard conditional LDMs without our proposed modules. As shown in the first row, this baseline produces legible but blurry text with poor style fidelity, achieving an FID of 16.73. While Base+SAQ dramatically enhances visual quality, causing the FID to plummet to 12.47. Next, Base+SAQ+$L_{SCE}$ helps improve HWD score from 0.85 to 0.84, demonstrating that this loss successfully enforces a more discriminative style space. Finally, Base+SAQ+$L_{SCE}$+$L_{LatentPCE}$ provides another major boost in local details, bringing FID down to an excellent 10.20 and HWD to 0.74. The visual results prove that all components work in synergy to achieve our final SOTA performance.

\begin{table}[tbh]
    \centering
    \vspace{-0.1in}
    \scalebox{0.5}{
        \begin{tabular}{p{0.07\linewidth}p{0.07\linewidth}p{0.07\linewidth}p{0.09\linewidth}||c c |c|c}
         %\toprule
         \hline
            %\cmidrule{1-8}
        ~ & ~ & ~ & ~ &  \multicolumn{2}{c|}{Style images} & FID $\downarrow$ & HWD $\downarrow$ \\
         \cline{5-6}
         
      Base  & {$SAQ$}   & {$L_{SCE}$} & $L_{PCE}$ &
         
         \includegraphics[scale=0.2, margin=0pt 1ex 0pt 1ex, valign=c]{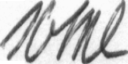}
            & \includegraphics[scale=0.2, margin=0pt 1ex 0pt 1ex, valign=c]{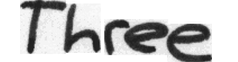} & ~   \\
         % \hline
            \midrule
            % \hline
    
          $\checkmark$&     &    &     & \includegraphics[scale=0.2, margin=0pt 1ex 0pt 1ex, valign=c]{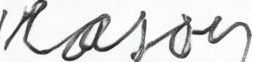} & \includegraphics[scale=0.2, margin=0pt 1ex 0pt 1ex, valign=c]{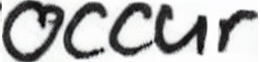} &  16.73 & 0.87  \\
       $\checkmark$&   $\checkmark$    &   &    & \includegraphics[scale=0.2, margin=0pt 1ex 0pt 1ex, valign=c]{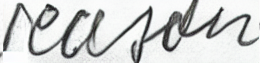} & \includegraphics[scale=0.2, margin=0pt 1ex 0pt 1ex, valign=c]{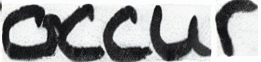} &  12.47 & 0.85 \\
      $\checkmark$&      $\checkmark$  &  $\checkmark$  &   & \includegraphics[scale=0.2, margin=0pt 1ex 0pt 1ex, valign=c]{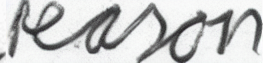} & \includegraphics[scale=0.2, margin=0pt 1ex 0pt 1ex, valign=c]{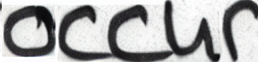} &  12.55 & 0.84
            \\
       $\checkmark$&    $\checkmark$     &  $\checkmark$  & $\checkmark$ & \includegraphics[scale=0.2, margin=0pt 1ex 0pt 1ex, valign=c]{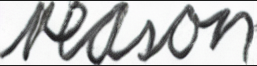} & \includegraphics[scale=0.2, margin=0pt 1ex 0pt 1ex, valign=c]{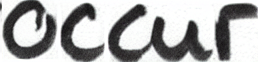} & \textbf{10.20} & \textbf{0.74} \\
             
            %\bottomrule
            \hline
        \end{tabular}
    }
    \caption{Contribution of our SAQ, $L_{SCE}$, and $L_{LatentPCE}$ to the baseline model in terms of FID and HWD score.}
    \label{tab:componentablation}
    % \vspace{-0.1in}
\end{table}

% \begin{figure}[htbp]
%     \centering 
% \includegraphics[width=0.7\linewidth]{figures/mainfigure-pcevscosine.drawio (1).pdf}
%     \caption{Qualitative comparisons between $L_{LatentPCE}$ and $L_{cosine}$}
%     \label{fig:pcevscosine}
%     % \vspace{-0.1in}
% \end{figure}
\vspace{-0.1in}
\subsubsection{Effectiveness of $L_{LatentPCE}$}
To validate our proposed objective, we compare it against alternative auxiliary objectives, including a global cosine similarity loss ($L_{cosine}$), LPIPS~\cite{zhang2018unreasonable}, and patch-wise L2 ($L_{PatchL2}$). As shown in Tab.~\ref{tab:pcevscosinetab}, adding $L_{LatentPCE}$ yields the lowest FID (14.01), outperforming the baseline (16.73) and other variants (e.g., $L_{cosine}$ at 21.06, LPIPS at 14.05, $L_{PatchL2}$ at 15.10), confirming its superior local detail sharpening. This stems from maximizing mutual information across multi-scale patches, unlike the $L_{cosine}$, which optimizes overall feature similarity without local constraints, or LPIPS and $L_{PatchL2}$, which, despite LPIPS leveraging VGG-pretrained features, both rely on L2 distances to minimize patch discrepancies but fail to encourage discriminative representations.

\subsection{Generalization to Other Languages}
\begin{figure}[tbhp]
    \centering 
\includegraphics[width=0.8\linewidth]{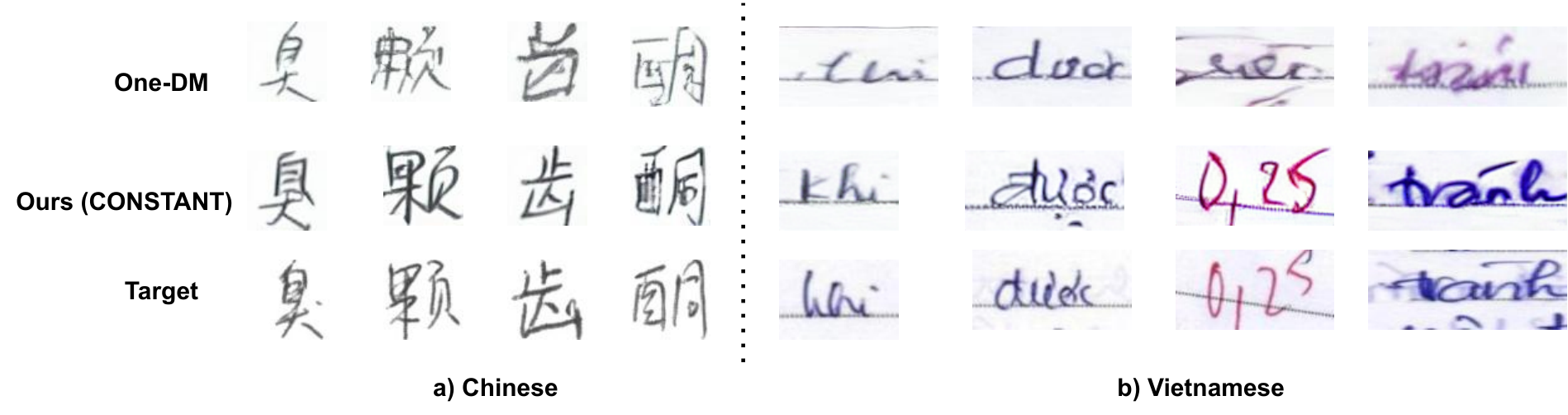}
    \caption{Comparisons between CONSTANT and One-DM~\cite{dai2025one} on Chinese and Vietnamese scripts.}
    \label{fig:otherlang}
    \vspace{-0.2in}
\end{figure}

To assess the adaptability and effectiveness of our proposed method, we extended our experiments beyond English to include languages with distinct writing systems. We consider performing experiments on Chinese language and Vietnamese language. For Chinese, we use CASIA database~\cite{yin2013icdar} containing 60 writers and 3755 Chinese characters for each writer. For Vietnamese, we construct a novel dataset with a variety of styles and diverse backgrounds, named ViHTGen. The dataset consists of more than 50000 Vietnamese handwriting images collected from the university's exam~\footnote{More information about our ViHTGen is presented in the Appendix}. We then compare our model against One-DM~\cite{dai2025one} method. Tab.~\ref{tab:chinesevietnam} shows that our method outperforms One-DM on both metrics with over 10\% improvement on HWD scores, indicating that our method adapts to style better than One-DM. From Fig.~\ref{fig:otherlang}a, we observe that our method effectively adapts to complex styles, such as varying stroke density, while preserving the content of the characters. In contrast, One-DM tends to generate blurry details and struggles to produce sufficient strokes in its handwriting. For Vietnamese scripts from Fig.~\ref{fig:otherlang}b, our method can effectively handle complex backgrounds with difficult character shapes and diverse colors, even for numeric characters. Meanwhile, One-DM fails to generate coherent results, with the foreground tending to blend with the background, indicating its failure to extract discriminative style concepts. \textit{More quantitative results with other methods and additional qualitative results are presented in the Appendix.}

\begin{table}[htbp] 
\centering
    % \vspace{-0.1in}
    \scalebox{0.65}{
        \begin{tabular}{c||*{2}{c}|*{2}{c}}
        \toprule
                        & \multicolumn{2}{c}{Chinese} & \multicolumn{2}{c}{Vietnamese}\\
                        Method  & HWD $\downarrow$       & FID $\downarrow$          & HWD $\downarrow$       & FID $\downarrow$          \\
                        \midrule
                        One-DM~\cite{dai2025one} & 0.48 & 22.97 & 1.08 & 22.53 \\
                        \midrule
                        \textbf{Our} & \textbf{0.37} & \textbf{22.74} & \textbf{0.83} & \textbf{18.81} \\
                        \bottomrule
        \end{tabular}
    }
\caption{Quantitative comparisons with One-DM on Chinese and Vietnamese scripts in terms of FID and HWD.}
    \label{tab:chinesevietnam}
    \vspace{-0.2in}
\end{table}

% \vspace{-0.15in}
\subsection{Analysis}
\subsubsection{Style Extraction Analysis}
% \vspace{-0.1in}
% \begin{figure}[htbp]
%     \centering 
% \includegraphics[width=0.8\linewidth]{figures/all-umapmain.drawio.pdf}
%     \caption{Style embedding space between real and generated data by our method}
%     \label{fig:umap}
%     % \vspace{-0.1in}
% \end{figure}

\begin{figure}[htbp]
  \centering
    \includegraphics[width=0.7\linewidth]{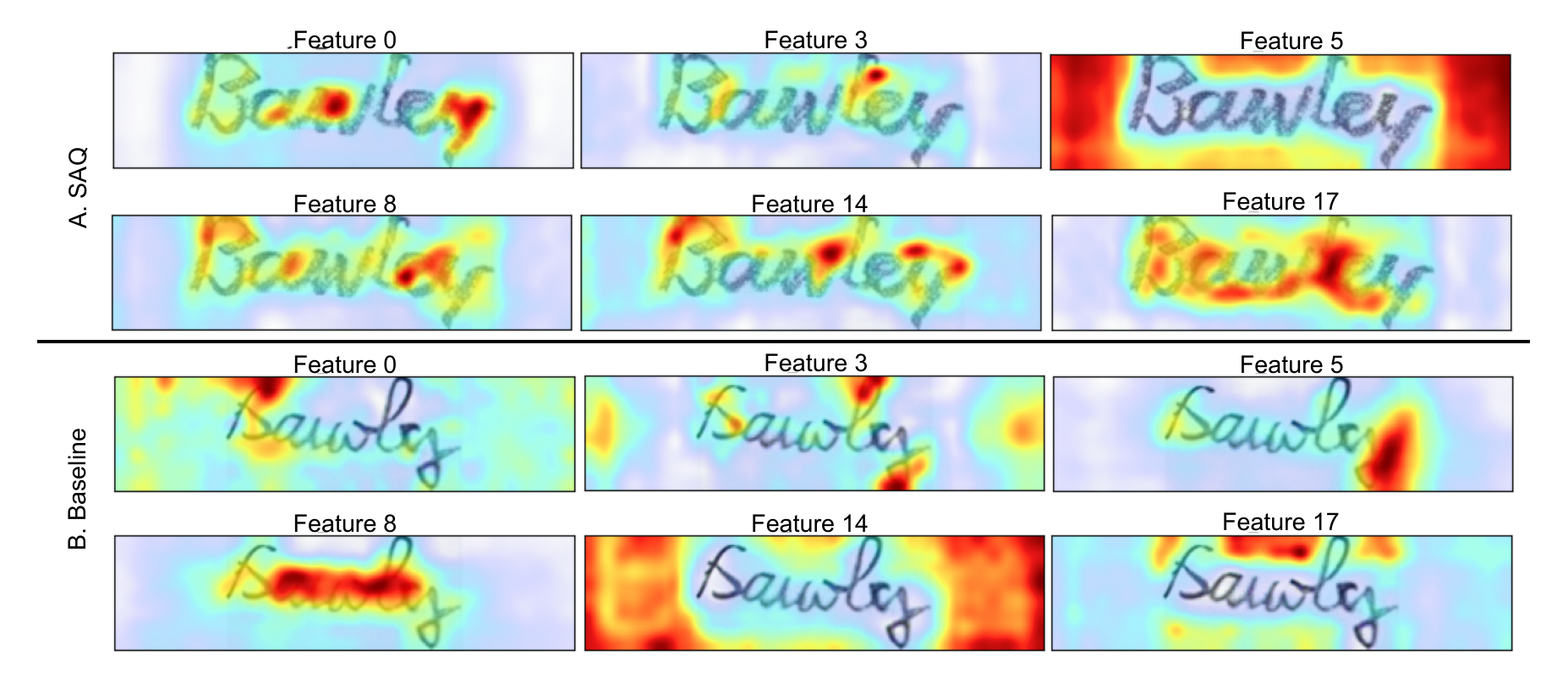}
    \caption{Style cross-attention map visualization between our CONSTANT and baseline model.}
    \label{fig:crossattn}
    \vspace{-0.1in}
\end{figure}

To better understand the behavior of our SAQ module, we conduct an interpretability analysis between SAQ and a baseline Inception-V3 (Sec.~\ref{sec:saq}) by visualizing the cross-attention scores between style features and generated pixels, inspired by~\cite{tang2022daam}. The resulting attention maps in Fig.~\ref{fig:crossattn} reveal a clear distinction: our SAQ module (A) produces sharp, localized maps where each feature corresponds to a specific character stroke. In contrast, the baseline's attention (B) is diffuse and unfocused, demonstrating that our SAQ module successfully learns to disentangle style into distinct, meaningful concepts.

\subsubsection{Domain Generalization Analysis}
\begin{figure}[H]
    \centering 
\includegraphics[width=0.8\linewidth]{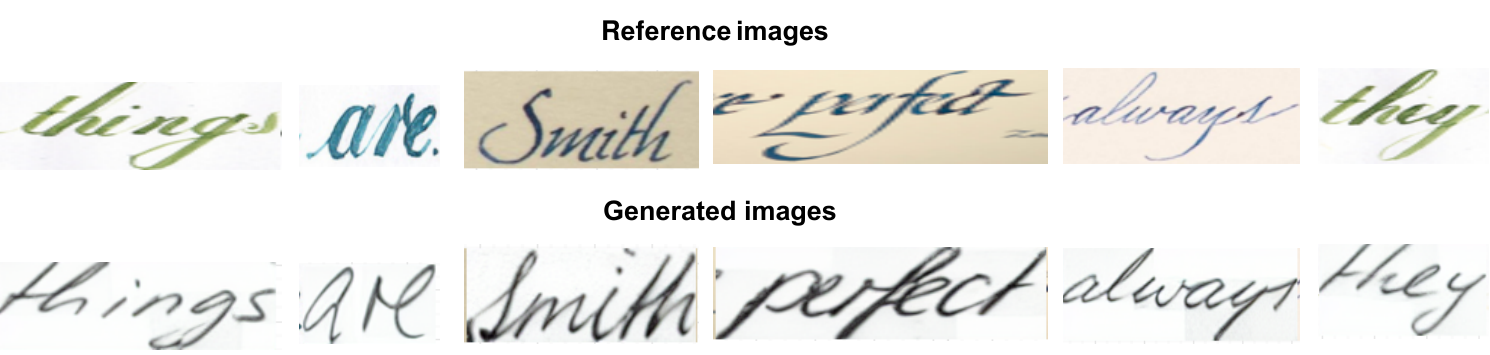}
    \caption{Real images versus generated images using IAM-trained model on IMGUR5K references.}
    \label{fig:domainadapt1}
    % \vspace{-0.1in}
\end{figure}

\begin{figure}[htpb]
\vspace{-0.1in}
\centering
\begin{subfigure}[b]{0.4\linewidth}
    \includegraphics[width=\linewidth]{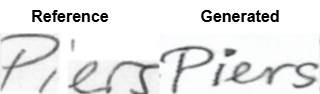}
    \caption{Example 1}
\end{subfigure}
\hskip-0.1\baselineskip
\begin{subfigure}[b]{0.5\linewidth}
    \includegraphics[width=\linewidth]{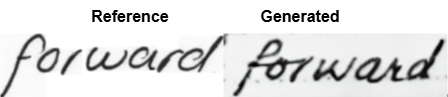}
    \caption{Example 2}
\end{subfigure}
\caption{Performing domain adaptation on IMGUR5K model using IAM reference images.}
\label{fig:iamonimgur5k}
\vspace{-0.2in}
\end{figure}

To investigate our model's cross-domain generalization, we conducted two challenging experiments. First, we evaluated the robustness to out-of-domain features by testing an IAM-trained model with style references from IMGUR5K. As shown in Fig.~\ref{fig:domainadapt1}, the model successfully preserves the learned structural style while correctly ignoring unseen concepts like ink color. Second, we evaluated whether our model, trained on a complex domain, could accurately generate clean styles by using an IMGUR5K-trained model with clean IAM references. Fig.~\ref{fig:iamonimgur5k} shows it can accurately generate clean styles without \say{hallucinating} the complex styles from its training domain. Together, these scenarios demonstrate our method's strong ability to generalize across domains with different characteristics.

% \begin{table}[htbp] \centering
%     \caption{Performing domain adaption on IMGUR5K model using IAM reference images}
%     \label{tab:iamonimgur5k}
%     \includegraphics[width=0.5\linewidth,height=0.065\textwidth]{figures/adapt2.png}
%     \includegraphics[width=0.5\linewidth,height=0.065\textwidth]{figures/adapt2-2.png}
%     % \raisebox{0.9\height}{
%     % \resizebox{0.15\textwidth}{!}{
%     % \large
%     % \begin{tabular}{cc}
%     %     \toprule
%     %     Method &  FID $\downarrow$ \\
%     %     \midrule
%     %     One-DM~\cite{dai2025one} & 77.91 \\ 
%     %     \textbf{CONSTANT}  & \textbf{25.29} \\
%     %     \bottomrule
%     % \end{tabular}
%     % }}
%     \\
%     \vspace{-0.2em}
%     \makebox[0.11\textwidth]{\footnotesize (a) Example 1} 
%     \makebox[0.16\textwidth]{\footnotesize (b) Example 2}
%     % \makebox[0.19\textwidth]{}
%     \\
% \end{table}

\section{Conclusion}
In this paper, we introduce CONSTANT, a novel diffusion-based method that significantly improves one-shot HTG. Our approach uses a Style-Aware Quantization (SAQ) module, enhanced by a contrastive loss ($L_{SCE}$), to robustly capture core style concepts from a single reference image through the innovative application of VQ. To address common issues like blurriness, we also incorporate a multiscale patch-level ($L_{LatentPCE}$) objective that refines fine-grained local details. Extensive experiments on multilingual datasets, including English, Chinese, and our new ViHTGen dataset, confirm our method's superior performance across visual quality, style fidelity, readability, and cross-domain generalization.

\clearpage
\maketitlesupplementary
\appendix

% Resetting counters for Supplementary labeling (e.g., Figure S1, Table S1)
\setcounter{page}{1}
\setcounter{equation}{0}
\setcounter{figure}{0}
\setcounter{table}{0}

\section{Introduction}
In this supplementary material, we provide more details about Latent Diffusion Models in Sec.~\ref{sec:ldm}. In Sec.~\ref{sec:users}, we conduct user studies comparing SOTA methods to evaluate visual and style imitation quality. Sec.~\ref{sec:vihtgen} provides detailed information about our ViHTGen dataset. We also present the implementation details more clearly in Sec.~\ref{sec:imple}. We provide more detail about our evaluation metrics in Sec.~\ref{sec:metric}. Next, we show the experimental results on the IIIT-English-Word dataset in Sec.~\ref{sec:iiit}. Sec.~\ref{sec:cross} shows more results between our method and other SOTA methods on multi-language generalization. In Sec.~\ref{sec:ocr}, we conduct an experiment to evaluate the effectiveness of our method on improving handwritten text recognition performance. We also perform an ablation study on the impact of codebook embedding length on the performance of our model in Sec.~\ref{sec:codebook}. We provide more detail about the efficiency aspect, including our training cost and inference time, in Sec.~\ref{sec:cost}. A discussion about limitations and future work is presented in Sec.~\ref{sec:limit}. Finally, we provide more qualitative results of our model on multiple datasets.

\section{More Details about LDMs}
\label{sec:ldm}
Diffusion models (DMs) represent a significant advance in generative modeling, often surpassing GANs in various tasks. Starting with DDPM~\cite{ddpm_nips_2020}, which uses iterative denoising to generate samples, numerous improvements have enhanced quality and control~\cite{ddim_iclr_2021,dhariwal2021diffusion,rombach2022high}. Techniques like classifier-free guidance (CFG)~\cite{ho2022classifier} and multimodal conditioning as seen in GLIDE~\cite{nichol2022glide} have further boosted performance, especially in text-to-image generation. \\
Our diffusion model for handwritten generation is inspired by LDMs~\cite{rombach2022high}, which enables the sampling process to occur in the latent space by using a pretrained VAE to compress handwritten image X to a 4-D latent space representation $z \in \mathbb{R}^{4 \times W / 8 \times H / 8}$. Similar to DDPM~\cite{ddpm_nips_2020}, the forward process considered a Markov chain of length T consists of gradually adding Gaussian noise to the clean latent representation $\mathbf{z}_0 \sim q\left(\mathbf{z}_0\right)$ until it becomes pure noise $p\left(\mathbf{z}_T\right)=\mathcal{N}\left(\mathbf{z}_T ; \mathbf{0}, \mathrm{I} \right)$, with each step defined by a transition probability $q(z_t|z_{t-1})$. \\
The reverse process aims to learn to undo this noise addition, generating a sample from pure noise by training a Unet model~\cite{ronneberger2015u} $\boldsymbol{\epsilon}_\theta\left(\boldsymbol{z}_t, t, \boldsymbol{C}, \boldsymbol{X}_s\right)$ to predict the added noise using a mean-squared error loss:
\begin{equation}
\resizebox{.91\linewidth}{!}{$
            \displaystyle
    L_{\text {denoising}}=E_{t \sim[1, T], z_0 \sim q\left(z_0\right), \epsilon \sim \mathcal{N}(0, \mathbf{I})}\left[\left\|\epsilon-\boldsymbol{\epsilon}_\theta\left(\boldsymbol{z}_t, t, \boldsymbol{C}, \boldsymbol{X}_s\right)\right\|^2\right]
    $}
    % \vspace{-0.1in}
\end{equation}

\section{User Studies}
\label{sec:users}
We perform user studies, including a User Preference Study and User Plausibility Study to better evaluate our methods compared to the others, including One-DM~\cite{dai2025one}, HiGAN+~\cite{HiGANplus}, HWT~\cite{HWT_2021_ICCV}, VATr~\cite{VATr_2023_CVPR}, HiGAN~\cite{HiGan_Wang_2021}, using TypeForm \footnote{https://www.typeform.com/} survey platform.

\subsection{User Preference Study}
\begin{figure}[htbp]
  \centering
    \includegraphics[width=\linewidth]{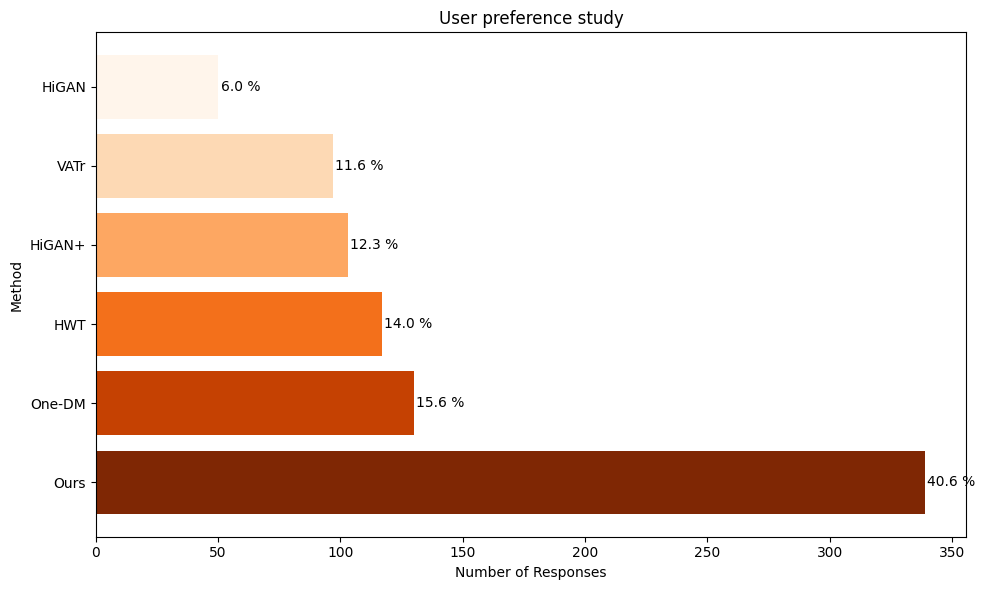}
    \caption{User preference study results}
    \label{fig:preference}
\end{figure}

We perform sampling of a list of 30 text contents using an OOV corpus from IAM~\cite{marti2002iam} dataset and generating 30 images for each method. The test was conducted on 28 participants and received a total of 840 responses. Each participant was asked to choose which image in the given list was the most similar to the real image. As shown in Fig.~\ref{fig:preference}, our method receives the most responses from all participants, with more than 40.6\% responses.

\subsection{User Plausibility Study}
\begin{figure}[htbp]
  \centering
    \includegraphics[width=\linewidth]{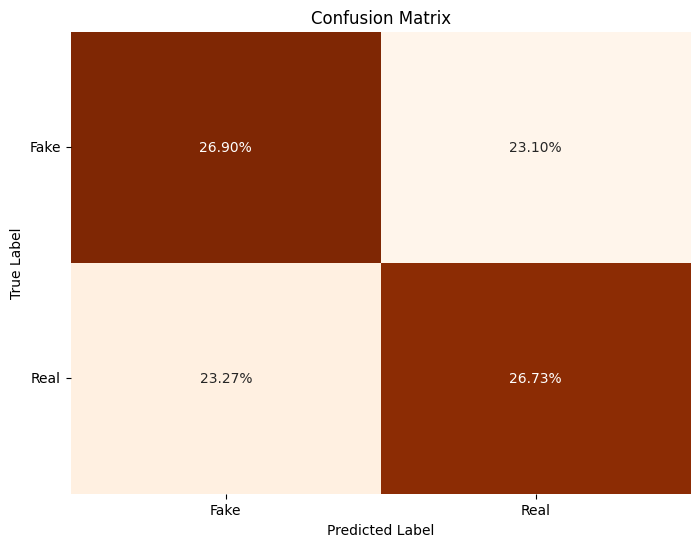}
    \caption{User plausibility study results}
    \label{fig:plausi}
\end{figure}

We perform experiment to study whether our method's generated images are indistinguishable from real images. For each question, we ask participants to identify 3 real images in total of 6 images by first showing them 6 examples of real images from the same writers. The study received a total of 1680 responses from a total of 28 participants. The result is shown in Fig.~\ref{fig:plausi} with the \textbf{accuracy of 53.8\%}, indicating it is close to random classification.

\section{ViHTGen Dataset}
\label{sec:vihtgen}
To mimic real-world scenarios, we constructed the \textbf{ViHTGen} dataset, featuring diverse handwriting styles on complex backgrounds. The dataset was sourced from over 300 university final exam scripts written by more than 200 individuals. We employed a rigorous semi-automatic annotation pipeline to ensure high-quality labels. First, word instance bounding boxes were automatically extracted using the Google Cloud Vision API\footnote{https://cloud.google.com/vision/docs} and then manually verified. Subsequently, a state-of-the-art Vietnamese vision-language model VinTernVL-1B~\cite{doan2024vintern} performed OCR on each word, with all transcriptions being manually checked and corrected to guarantee label accuracy.

\begin{figure*}[thbp]
  \centering
    \includegraphics[width=0.8\linewidth]{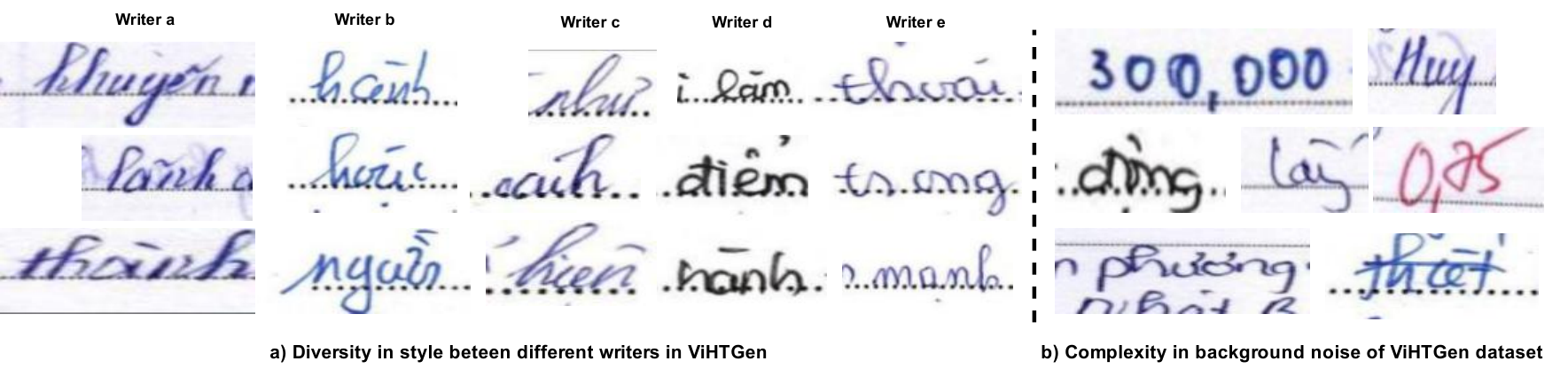}
    \caption{An overview of our ViHTGen dataset. a) The diversity of handwritten style between different writers. b) The complexity of our dataset in both background and stroke shape.}
    \label{fig:vihtgen}
\end{figure*}

\begin{figure}[htbp] % Or h!, t!, b! for float placement
    \centering % Center the two subfigures within the column

    \begin{subfigure}[b]{0.48\columnwidth} % [b] for bottom alignment, adjust width
        \centering
        \includegraphics[width=\textwidth]{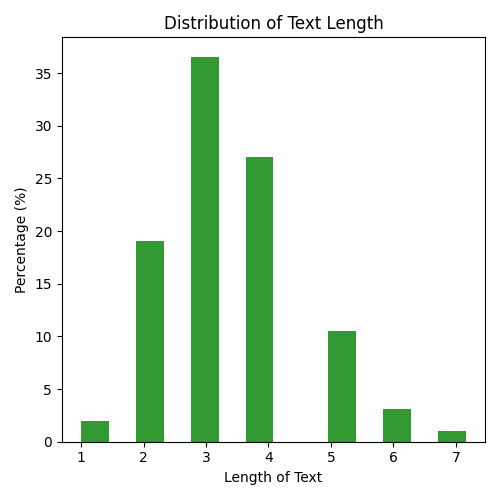} % image1.png should exist
        \caption{The distribution of text length}
        \label{fig:subfig1}
    \end{subfigure}
    \hfill % Adds horizontal space, pushing subfigures apart
    \begin{subfigure}[b]{0.48\columnwidth} % Adjust width
        \centering
        \includegraphics[width=\textwidth]{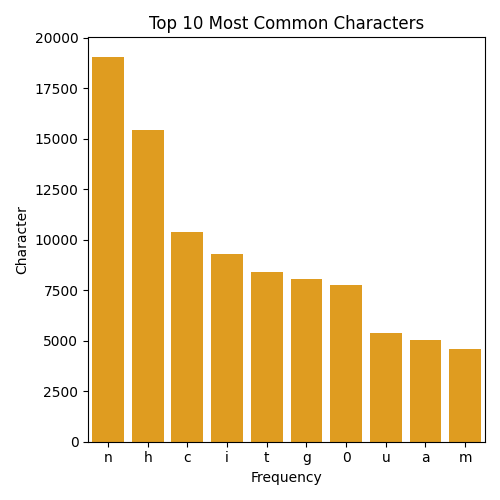} % image2.png should exist
        \caption{Most frequent character}
        \label{fig:subfig2}
    \end{subfigure}

    \caption{Statistical analysis for the ViHTGen dataset. (a) The percentage of images with different text lengths. (b) Frequency distribution of top-10 characters in the dataset}
    \label{fig:statistic}
\end{figure}

After filtering, the final dataset contains over 50,000 word-level images, which we split into a training set of 42,000 and a test set of over 8,000 images. As shown in Fig.~\ref{fig:statistic}, we also performed a statistical analysis of text length and character frequency. The resulting dataset is a challenging benchmark for HTG models, featuring a wide variety of stroke styles and complex, noisy backgrounds, as illustrated in Fig.~\ref{fig:vihtgen}. Tab.~\ref{tab:comaprevi} also shows the differences between our dataset and the standard IAM dataset on many aspects. To encourage reproducibility and foster future research, the ViHTGen dataset will be made publicly available upon publication of this work.

\begin{table}[htbp]
    \centering
    \begin{tabular}{|p{2.5cm}|c|c|}
    \hline
        Statistic & ViHTGen (ours) & IAM \\ \hline
        Language & Vietnamese & English  \\ \hline
        \# Writers & 223 & 500 \\ \hline
        \# Word Instances & 50000+ & 62857 \\ \hline
        \# Unique words & 4644 & 3332 \\ \hline
        Style Complexity (Slant, Ink color, Stroke width, character shape) & High & Medium \\ \hline
        Background complexity & High & Low \\ \hline
        Image source & University Exam & English Sentences\\ \hline
    \end{tabular}
    \caption{Comparison between our ViHTGen and IAM dataset.}
    \label{tab:comaprevi}
\end{table}

\section{More about Implementation Details}
\label{sec:imple}
\subsection{Model Architecture Details}
Our model includes three basic components: 1) a Latent Diffusion-based model, 2) a SAQ module for style feature extraction, and 3) a text encoder module.

\begin{itemize}
\item \textbf{Latent Diffusion-based Models}: We follow the architecture of LDMs~\cite{rombach2022high}, which uses a U-Net~\cite{ronneberger2015u} model that includes a ResNet block followed by a Spatial Transformer block to combine information between context features (here, the style and textual features) and the input features. We follow WordStylist~\cite{nikolaidou2023wordstylist} to reduce the number of ResNet blocks to reduce training time, while the number of heads in the Transformer block is set to 4, and the feature dimension is set to 512.
\item \textbf{SAQ module}: Our proposed SAQ module includes three basic parts: an Inception-V3 backbone, a codebook embedding $E$, and an AttentionPool fusion module. We chose Inception-V3 for its effectiveness in extracting multiscale features, as is also done in style transfer~\cite{ghiasi2017exploring}. The output dimension of Inception-V3 is 768. Since we use a hybrid solution that combines discrete and continuous features as described in the main paper, the output dimension of features $\hat F$ in SAQ is 1536. Finally, the AttentionPool module, for which we were inspired by the CLIP framework~\cite{radford2021learning}, is used to better fuse information from both discrete and continuous features through a self-attention operation. The final output dimension is projected to 512 through a linear layer before being passed to the LDM.
\item \textbf{Text encoder module}: The text encoder is a 3-layer Transformer block; each block consists of an MLP block and a self-attention block with a dimension of 512.
\end{itemize}

\subsection{$L_{LatentPCE}$ Implementation Details}
We design our $L_{LatentPCE}$ objective as a multi-scale contrastive loss that operates at three distinct scales. At each scale, we extract up to 256 patches of sizes $2 \times 2$, $4 \times 4$, and $8 \times 8$, respectively. These extracted patches are then flattened and projected into a 256-dimensional embedding space using a shallow MLP. The final $L_{LatentPCE}$ is computed as the average of the contrastive losses from each of the three scales.

\section{More Details About Evaluation Metrics}
\label{sec:metric}
\subsection{OCR-based Metric}
To evaluate the readability of the generated images, we follow the setup from~\cite{nikolaidou2024rethinking}. The core idea is to train an Optical Character Recognition (OCR) model on the generated images and then test its performance on real images. A successful HTG model should produce samples that enable the OCR model to achieve a low Word Error Rate (WER) on real data. We use the sequence-to-sequence OCR architecture from~\cite{kass2022attentionhtr}. For the evaluation, we first train the OCR model on images generated for the IAM test set. The model is trained for 200,000 iterations with a batch size of 128. After training, we test the OCR model on the real IAM test set and calculate the WER.

\subsection{Writer Classification Metric}
Besides evaluating readability, we also assess the style imitation ability using a writer classifier model. This evaluation strategy has been used in previous works~\cite{nikolaidou2023wordstylist, nikolaidou2024diffusionpen}. In our work, we use a ResNet18 model pre-trained on ImageNet as the base architecture. To train the classifier, we split the real IAM test set into an 80/20 ratio for training and validation. The trained model is then used to evaluate the writer classification accuracy on the generated version of IAM test set, which serves as a measure of our HTG model's style imitation capability. 

\section{Experiments on IIIT-English-Word Dataset}
\label{sec:iiit}

\begin{figure*}[thbp]
  \centering
    \includegraphics[width=\linewidth]{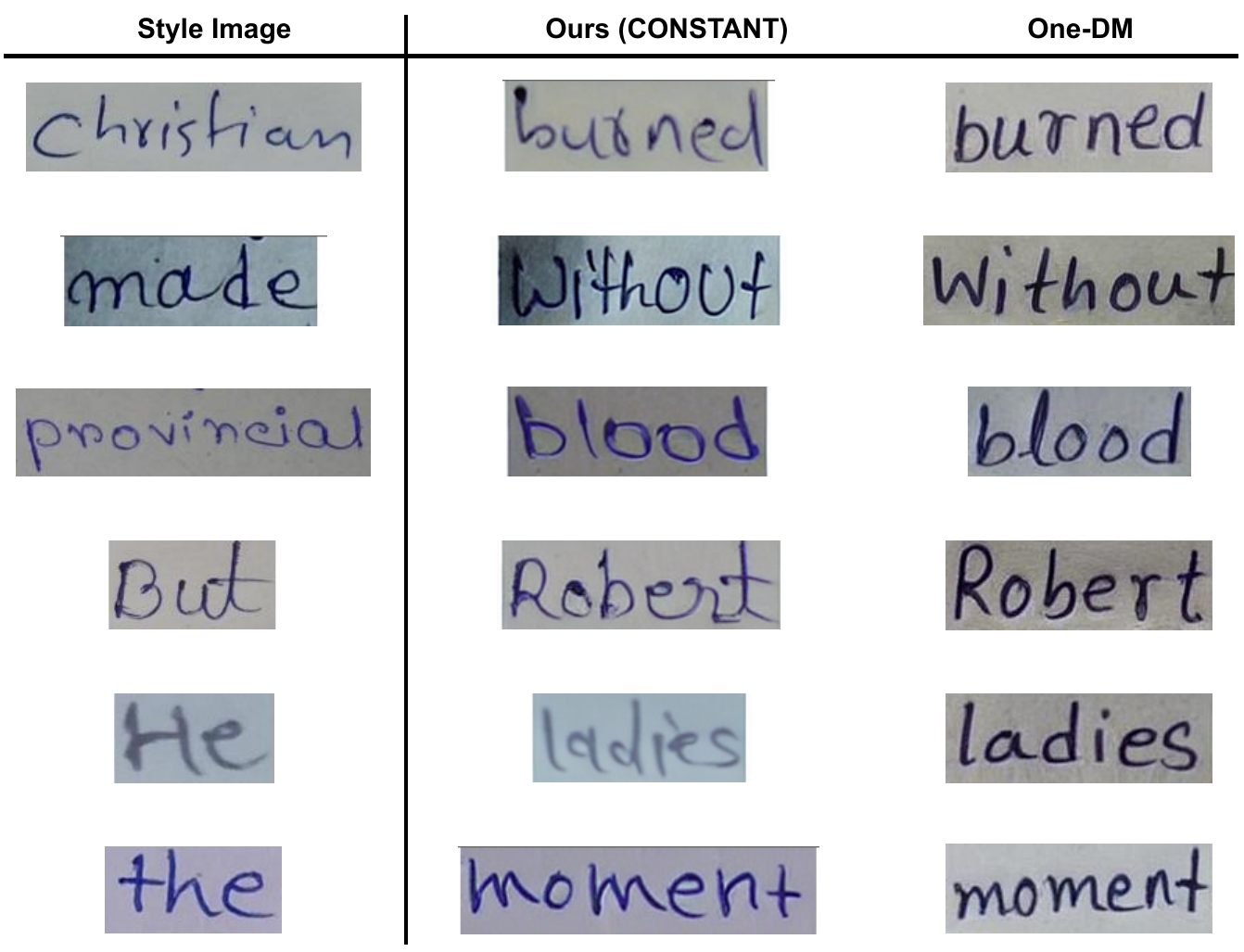}
    \caption{Comparisons between our method and One-DM~\cite{dai2025one} on IIIT-English-Word.}
    \label{fig:iiitresutl}
\end{figure*}

We perform experiments on IIIT-English-Word dataset~\cite{mondal2024bridging} to compare the performance between our CONSTANT model and One-DM~\cite{dai2025one} in terms of FID and HWD score. The quantitative result show in Tab.~\ref{tab:IIIT} and the qualitative result show in Fig.~\ref{fig:iiitresutl} 

\begin{table}[tbhp]
    % \vspace{-0.2in}
    \begin{center}
    \scalebox{0.8}{
    \large
        \begin{tabular}{c||c|c}
        \toprule
        Method      & HWD  $\downarrow$   & FID  $\downarrow$      \\
        \midrule
        % HiGAN+ \cite{HiGANplus}      & 1.35 & 20.04   \\
        % HiGAN \cite{HiGan_Wang_2021}       &   1.55      & \underline{17.58}  \\
        % WordStylist \cite{nikolaidou2023wordstylist} & 1.16  & 24.25    \\
        One-DM \cite{dai2025one}     & 1.22  & 17.74     \\
        \midrule
        \textbf{Ours}    & \textbf{0.73}  & \textbf{10.22} \\
        \bottomrule
        \end{tabular}
    }
    \caption{Quantitative results on IIIT-English-Word test set.} 
    % \vspace{-0.1in}
    \label{tab:IIIT}
    \end{center}
\end{table}
% {figures/supp/update/mainfigure-IIITmore.drawio.pdf

\section{More Results on Multi-language Generalization}
\label{sec:cross}
\begin{figure}[htbp]
  \centering
    \includegraphics[width=\linewidth]{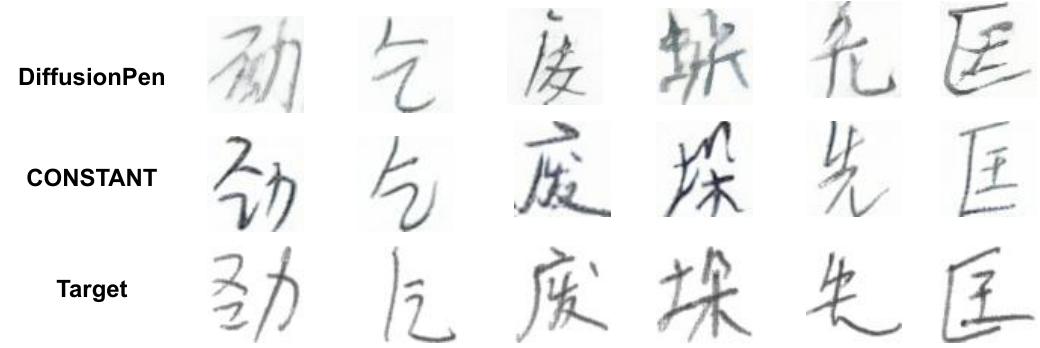}
    \caption{Qualitative results between CONSTANT and DiffusionPen~\cite{nikolaidou2024diffusionpen} in Chinese script.}
    \label{fig:penvsour}
\end{figure}

To better demonstrate the effectiveness of our method on other languages, we perform additional experiments comparing our method to a state-of-the-art (SOTA) competitor. In this experiment, we use DiffusionPen~\cite{nikolaidou2024diffusionpen} as the baseline model. Since DiffusionPen is a few-shot model, we train it in a one-shot setting to ensure a fair comparison. Similar to the evaluation against One-DM in the main paper, we compare our method with DiffusionPen on both Chinese and Vietnamese datasets. As shown in Tab.~\ref{tab:chinesevietnam_pen} and Fig.~\ref{fig:penvsour}, our method achieves better results on both languages in terms of visual quality and style adaptation ability compared to DiffusionPen. This also shows the robustness of our method in the one-shot setting, whereas the DiffusionPen model's performance is low when adapted to this setting (e.g., achieving a Chinese FID score of 54.07 compared to our 22.74).

\begin{table}[htbp] 
\centering
    % \vspace{-0.1in}
    \scalebox{0.65}{
        \begin{tabular}{c||*{2}{c}|*{2}{c}}
        \toprule
                        & \multicolumn{2}{c}{Chinese} & \multicolumn{2}{c}{Vietnamese}\\
                        Method  & HWD $\downarrow$       & FID $\downarrow$          & HWD $\downarrow$       & FID $\downarrow$          \\
                        \midrule
                        DiffusionPen~\cite{nikolaidou2024diffusionpen} & 0.57 & 54.07 & 1.05 & 23.59 \\
                        \midrule
                        \textbf{Our} & \textbf{0.37} & \textbf{22.74} & \textbf{0.83} & \textbf{18.81} \\
                        \bottomrule
        \end{tabular}
    }
\caption{Quantitative comparisons with One-DM on Chinese and Vietnamese scripts in terms of FID and HWD.}
    \label{tab:chinesevietnam_pen}
    \vspace{-0.2in}
\end{table}

\section{Handwritten Text Recognition Improvement}
\label{sec:ocr}
We conduct experiments to evaluate the improvement of HTR model when increasing the number of generated data in the training set. The experiments include training on the real IAM training set and gradually increasing the number of handwritten images generated using our methods by 100K each time.

\begin{table}[htpb]
    \centering
    \begin{tabular}{c||c}
    \toprule
        Data Source & Accuracy  \\
        \midrule
        Real data & 81.76 \\
        Real data + 100K & 82.13 \\
        \textbf{Real data + 200K} & \textbf{82.96} \\
        \bottomrule
    \end{tabular}
    \caption{Generated data helps improve OCR performance on IAM test set}
    \label{tab:ocrimprove}
\end{table}

\section{Ablation on Codebook Embedding Length}
\label{sec:codebook}
We conducted an ablation study to investigate the impact of the codebook embedding length in SAQ and to validate our hypothesis regarding its correlation with dataset complexity. The study involved varying the codebook size (K) across three datasets with distinct style complexities: the relatively simple IAM dataset and the more visually complex IMGUR5K and IIIT-English-Word datasets. We evaluated codebook sizes of K=512, K=1024, and K=2048. As shown in Tab.~\ref{tab:codebooklength}, the results align with our hypothesis. For the simple IAM dataset, the largest codebook (K=2048) slightly worsened performance (FID 14.10, HWD 0.86), which suggests overfitting. In contrast, for the more intricate styles of IMGUR5K and IIIT-English-Word, the larger K=2048 codebook yielded the best performance, enabling the model to capture a richer diversity of style features. These results effectively consolidate the hypothesis stated in our main paper.

\begin{table}[htbp] 
\centering
    % \vspace{-0.1in}
    \scalebox{0.65}{
        \begin{tabular}{c||*{2}{c}|*{2}{c}|*{2}{c}}
        \toprule
                        & \multicolumn{2}{c}{IAM} & \multicolumn{2}{c}{IMGUR5K}  & \multicolumn{2}{c}{IIIT-English-Word}\\
                        Codebook length  & HWD $\downarrow$       & FID $\downarrow$          & HWD $\downarrow$       & FID $\downarrow$ & HWD $\downarrow$       & FID $\downarrow$         \\
                        \midrule
                        512 & 0.86 & 13.80 & 1.02 & 13.37 & 0.75 & 14.02 \\
                        \midrule
                        1024 & \textbf{0.84} & \textbf{12.46} & 1.03 & 12.9  & 0.77 & 12.61 \\
                        \midrule
                        2048 & 0.86 & 14.10 & \textbf{0.99} & \textbf{11.48} & \textbf{0.73} & \textbf{10.22} \\
                        \bottomrule
        \end{tabular}
    }
\caption{Ablation Study About The Codebook Embedding Length in SAQ Module} 
    \label{tab:codebooklength}
    % \vspace{-0.2in}
    
\end{table}

% \begin{table}[tbhp]
% \caption{Ablation study about the codebook embedding length in SAQ module}
% \label{tab:codebooklength}
% \begin{center}
% \scalebox{0.8}{
% \small
% \begin{tabular}{c||cc|cc|cc}
% \toprule
% Codebook length & \multicolumn{2}{c|}{IAM} & \multicolumn{2}{c|}{IMGUR5K} & \multicolumn{2}{c}{IIIT-English-Word} \
% & HWD $\downarrow$ & FID $\downarrow$ & HWD $\downarrow$ & FID $\downarrow$ & HWD $\downarrow$ & FID $\downarrow$ \
% \midrule
% 512 & 0.86 & 13.80 & 0.92 & 15.20 & 0.88 & 14.50 \
% \midrule
% \textbf{1024} & \textbf{0.84} & \textbf{12.46} & 0.88 & 13.80 & \textbf{0.82} & \textbf{12.90} \
% \midrule
% 2048 & 0.86 & 14.10 & \textbf{0.85} & \textbf{13.20} & 0.84 & 13.60 \
% \bottomrule
% \end{tabular}
% }
% \end{center}
% \end{table}

\section{Effiency Analysis}
\label{sec:cost}
To provide further details on the training cost and efficiency of our method, we compare it with One-DM in terms of training cost. Tab.~\ref{tab:cost} provides detailed information on the training and inference costs between our method and the baseline One-DM model. Experiments for both methods were conducted on the same NVIDIA V100 machine.

\begin{table}[htbp]
    \centering
    \begin{tabular}{|p{2.5cm}|c|c|}
    \hline
        Criteria & CONSTANT & One-DM \\ \hline
        Model parameters & 124$\times 10^6$ & 185$ \times 10^6$  \\ \hline
        Training VRAM & 6.82GB & 18.66GB \\ \hline
        Inference time\footnotemark & 1.25 s/sample & 1.85 s/sample \\ \hline
        Inference VRAM & 2.1GB & 2.9GB \\ \hline
    \end{tabular}
    \caption{Comparison training and inference cost between CONSTANT and One-DM.}
    \label{tab:cost}
\end{table}
\footnotetext{Sampling taken under 50 steps}

\section{Limitation and Future Works}
\label{sec:limit}
\begin{figure}[htbp]
  \centering
    \includegraphics[width=\linewidth]{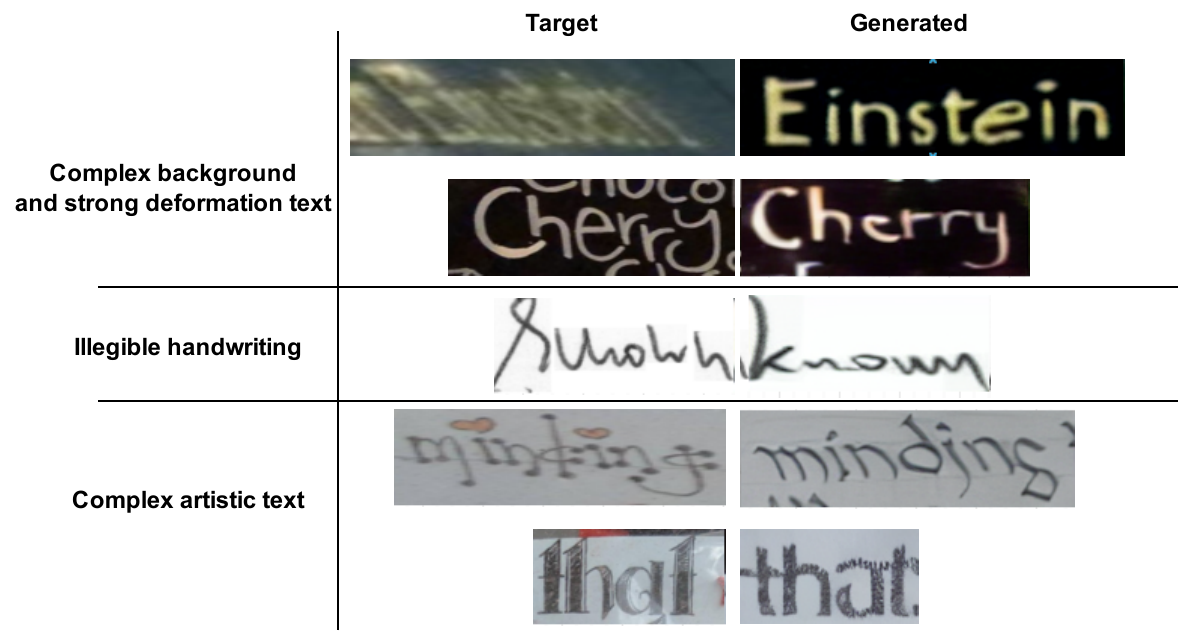}
    \caption{Visualization of some failure cases.}
    \label{fig:failure}
\end{figure}

Despite achieving state-of-the-art results, our method has limitations. The model's style extraction can be compromised when reference images are excessively blurry or feature highly complex backgrounds (Fig.~\ref{fig:failure}a). Similarly, for overly intricate or nearly illegible handwriting, our model may prioritize content readability over precise style imitation, as shown in Fig.~\ref{fig:failure}. Another limitation is that the codebook size is determined empirically, which may not be optimal for datasets with different style complexities. Future work will focus on improving style extraction for highly artistic text and exploring methods for generating longer lines of text.

\section{More Qualitative Results}
\label{sec:quali}
% \subsection{IAM}

\begin{figure*}[ht]
  \centering
    \includegraphics[width=\linewidth]{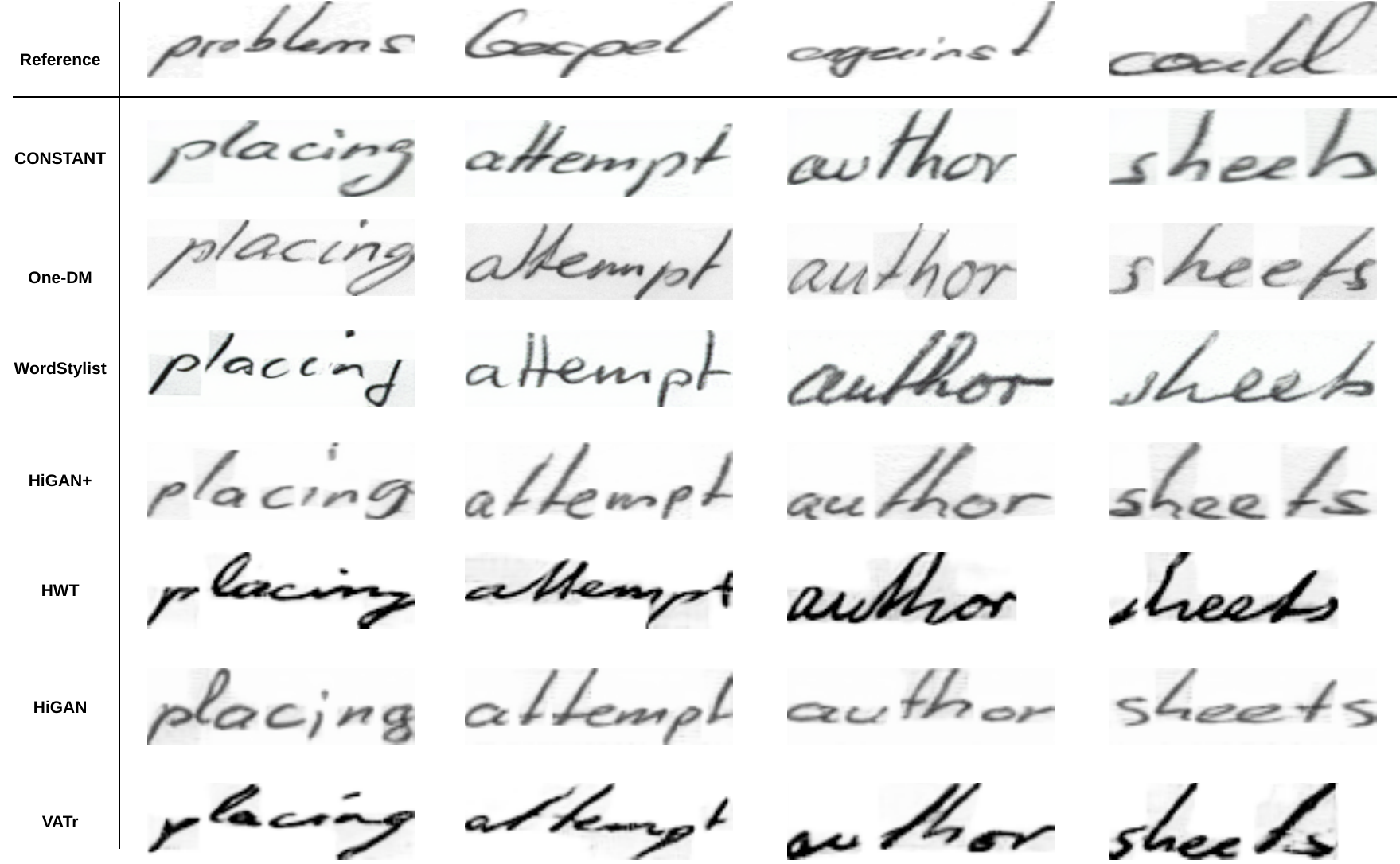}
    \caption{Visualization of arbitrary textual content between different methods on IAM test dataset.}
\end{figure*}

\begin{figure*}[h]
  \centering
    \includegraphics[width=\linewidth]{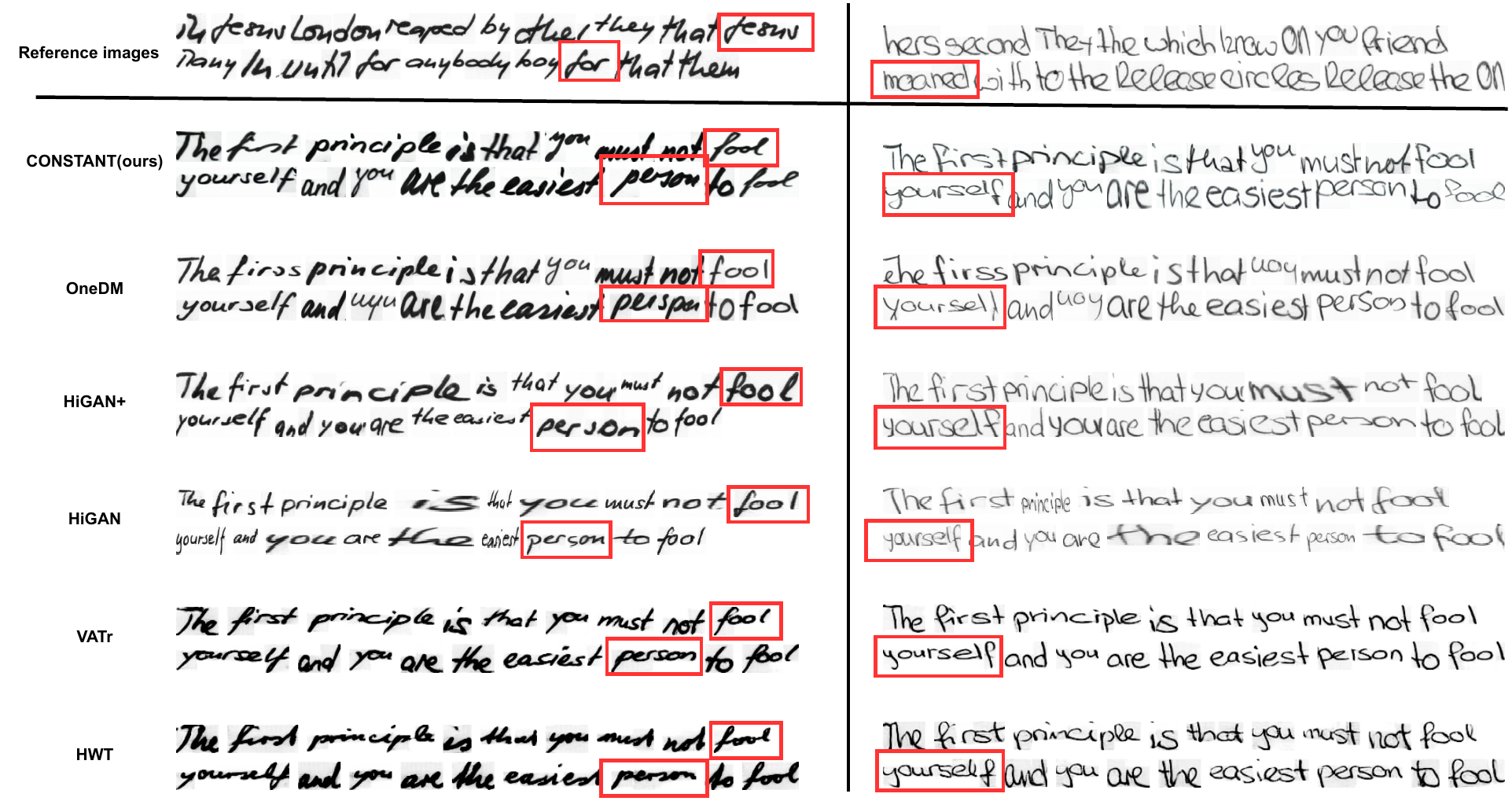}
    \caption{Visualization of our CONSTANT with other competitors, where \textcolor{red}{RED} box shows the style adaptability between different methods to the reference image.}
\end{figure*}

% \subsection{IMGUR5K}

\begin{figure*}[h]
  \centering
    \includegraphics[width=\linewidth]{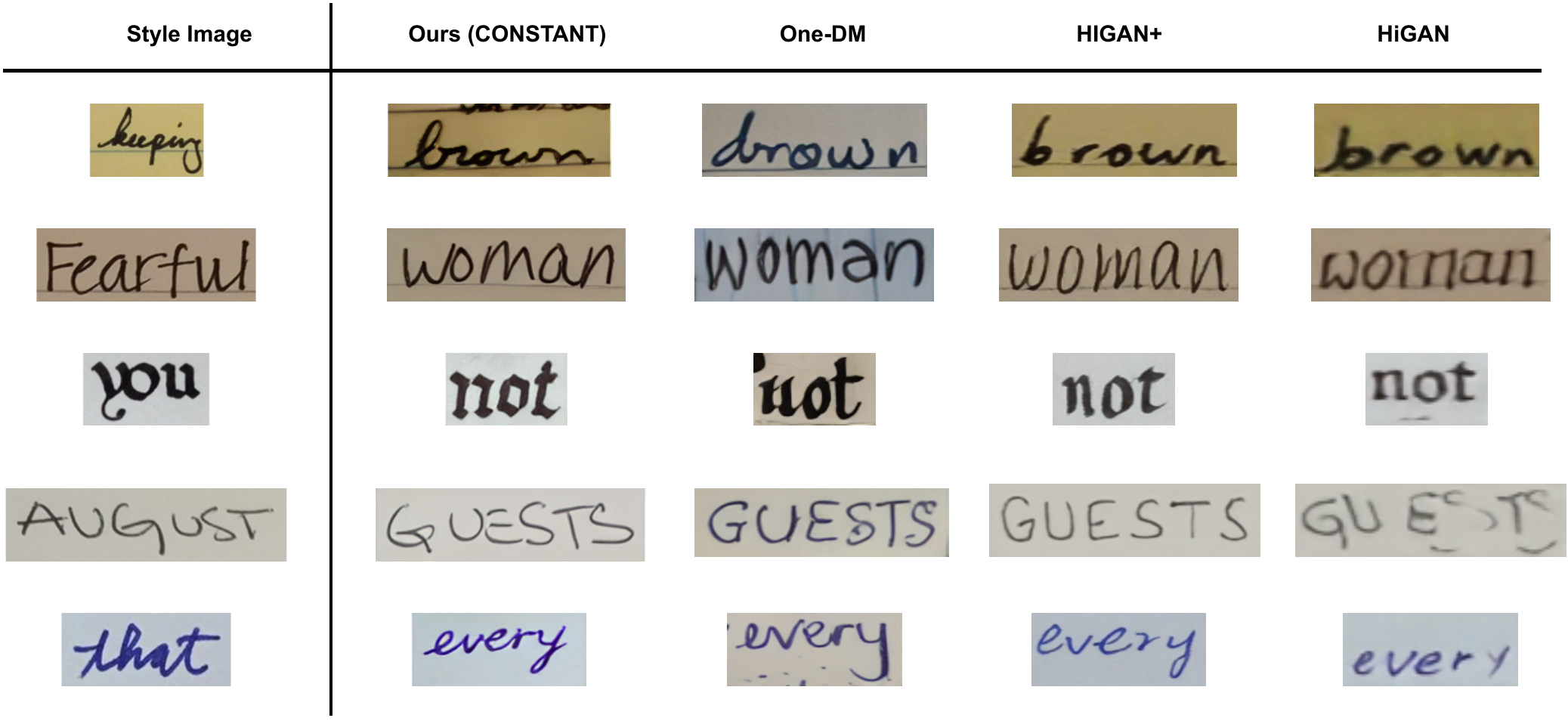}
    \caption{Other examples on IMGUR5K dataset.}
\end{figure*}

\begin{figure*}[h]
  \centering
    \includegraphics[width=\linewidth]{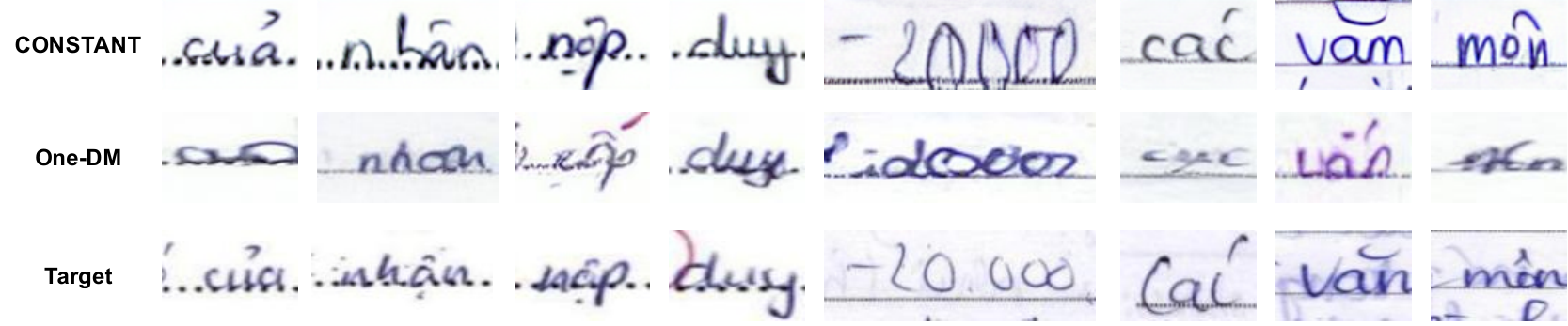}
    \caption{Other examples on ViHTGen dataset.}
\end{figure*}

% \subsection{IIIT-English-Word}

\newpage
\section*{Acknowledgments}
We acknowledge Ho Chi Minh City University of Technology (HCMUT), VNU-HCM for supporting this study.
{
    \small
    \bibliographystyle{ieeenat_fullname}
    \bibliography{main}
}

\end{document}